%% file: acl2019.tex
\PassOptionsToPackage{table}{xcolor}

\documentclass[11pt]{article}
\usepackage[hyperref]{acl2020}
\usepackage{times}
\usepackage{latexsym}

\usepackage[utf8]{inputenc} %
\usepackage[T1]{fontenc}    %
\usepackage{hyperref}       %
\usepackage{haldefs}
\usepackage{url}            %
\usepackage{booktabs}       %
\usepackage{amsfonts}       %
\usepackage{nicefrac}       %
\usepackage{microtype}      %
\usepackage{tipa}
\usepackage{bbm}
\usepackage{amssymb,amsmath,amsthm}

\usepackage{float}
\usepackage{xspace}
\usepackage{xparse}
\usepackage{amsthm}
\usepackage{tikz}
\usepackage{wrapfig}
\usepackage{array}
\usepackage[normalem]{ulem}

\hypersetup{colorlinks=true, linkcolor=darkpurple, citecolor=darkblue, filecolor=darkblue, urlcolor=darkblue}

\usepackage{url}

\theoremstyle{definition}
\newtheorem{theorem}{Theorem}

\newtheorem*{theorem*}{Theorem}
\newtheorem*{lemma*}{Lemma}

\aclfinalcopy %

\newcommand{\ourname}{\textsc{LeaQI}\xspace}
\newcommand{\ournamelong}{Learning to Query for Imitation\xspace}
\newcommand{\ournameipa}{\textipa{"li:,tSi:}}

\newcommand{\pistar}{\pi^\star}

\newcommand{\piref}{\pi^{\text{h}}}

\newcommand{\PExp}{\textsc{DAgger}\xspace}
\newcommand{\AExp}{\textsc{ActiveDAgger}\xspace}
\newcommand{\PRef}{\textsc{DAgger+Feat.}\xspace}
\newcommand{\ARef}{\textsc{ActiveDAgger+Feat.}\xspace}
\newcommand{\RandRef}{\textsc{LeaQI+NoisyHeur.}\xspace}
\newcommand{\OurNoAt}{\textsc{LeaQI+NoAT}\xspace}

\setlist{itemsep=0.25em}

\newcommand{\eps}{\epsilon}

\newcommand{\regret}{\textit{reg}}

\newcommand{\avgi}{\frac 1 N \sum_{i=1}^N}
\newcommand{\epspol}{\eps_{\text{pol-approx}}}
\newcommand{\epsdc}{\eps_{\text{dc-approx}}}
\newcommand{\regpol}{\regret_{\text{pol}}}

\newcommand{\err}{\textit{err}}
\newcommand{\Epi}{\Ep_{s \sim d_{\pi_i}}}

\newcount\Comments  %
\definecolor{darkgreen}{rgb}{0,0.5,0}
\definecolor{darkred}{rgb}{0.7,0,0}
\definecolor{teal}{rgb}{0.3,0.8,0.8}
\definecolor{purple}{rgb}{0.5,0.0,0.5}
\newcommand{\kibitz}[2]{\ifnum\Comments=1\textcolor{#1}{#2}\fi}

\newcommand\blacksout{\bgroup\markoverwith{\textcolor{black}{\rule[0.3ex]{10pt}{2pt}}}\ULon}

\title{Active Imitation Learning with Noisy Guidance}

\author{Kiant\'e Brantley\\
	University of Maryland\\
    \texttt{kdbrant@cs.umd.edu}\\ \And
    Amr Sharaf\\
    University of Maryland\\
      \texttt{amr@cs.umd.edu}\\ \And    
	Hal Daum\'e III\\
  	University of Maryland\\
  	Microsoft Research \\
  \texttt{me@hal3.name}}

\makeatletter
\DeclareRobustCommand\citepos{\begingroup\let\NAT@nmfmt\NAT@posfmt\NAT@swafalse\let\NAT@ctype\z@\NAT@partrue\@ifstar{\NAT@fulltrue\NAT@citetp}{\NAT@fullfalse\NAT@citetp}}
\let\NAT@orig@nmfmt\NAT@nmfmt
\def\NAT@posfmt#1{\NAT@orig@nmfmt{#1's}}
\makeatother

\date{}

\begin{document}
\maketitle

\begin{abstract}
  \input{abstract}

\end{abstract}

\section{Introduction} \label{sec:intro} \input{intro}

\section{Background and Related Work} \label{sec:related} \input{related}

\section{Our Approach: \ourname} \label{sec:method} \input{method}

\section{Experiments} \label{sec:applications} \input{applications}

\section{Discussion and Limitations} \label{sec:discussion} \input{discussion}

\section*{Acknowledgements}

We thank Rob Schapire, Chicheng Zhang, and the anonymous ACL  reviewers for very helpful comments and insights.  This
material is based upon work supported by the National Science Foundation under Grant No. 1618193 and an ACM SIGHPC/Intel Computational and Data Science Fellowship to KB. Any opinions, findings, and conclusions or recommendations expressed in this material are those of the author(s) and do not
necessarily reflect the views of the National Science Foundation nor of the ACM.

\bibliography{bibfile-shortname,bibfile,acl2019}
\bibliographystyle{acl_natbib}

\appendix
\newpage
\onecolumn
\begin{centering}
  \Large
  Supplementary Material For:\\
  Active Imitation Learing with Noisy Guidance\\
\end{centering}
\section{Experimental Details:} \label{sec:appendix_experiments} \input{appendix_experiments}

\end{document}

%% file: abstract.tex
Imitation learning algorithms provide state-of-the-art results on many structured prediction tasks by learning near-optimal search policies.
Such algorithms assume training-time access to an expert that can provide the optimal action at any queried state; unfortunately, the number of such queries is often prohibitive, frequently rendering these approaches impractical. 
To combat this query complexity, we consider an active learning setting in which the learning algorithm has additional access to a much cheaper \emph{noisy heuristic} that provides noisy guidance.
Our algorithm, \ourname, learns a \emph{difference classifier} that predicts when the expert is likely to disagree with the heuristic, and queries the expert only when necessary.
We apply \ourname to three sequence labeling tasks, demonstrating significantly fewer queries to the expert and comparable (or better) accuracies over a passive approach.

%% file: intro.tex
Structured prediction methods learn models to map inputs to complex outputs with internal dependencies, typically requiring a substantial amount of expert-labeled data.
To minimize annotation cost, we focus on a setting in which an expert provides labels for \emph{pieces} of the input, rather than the complete input (e.g., labeling at the level of words, not sentences).
A natural starting point for this is imitation learning-based ``learning to search'' approaches to structured prediction~\citep{daume09searn,ross11dagger,bengio-scheduled,leblond18searnn}.
In imitation learning, training proceeds by incrementally producing structured outputs on piece at a time and, at every step, asking the expert ``what would you do here?'' and learning to mimic that choice.
This interactive model comes at a substantial cost: the expert demonstrator must be continuously available and must be able to answer a potentially large number of queries.

\begin{figure*}[t]
  \hspace{-0.85cm}\includegraphics[width=1.06\linewidth]{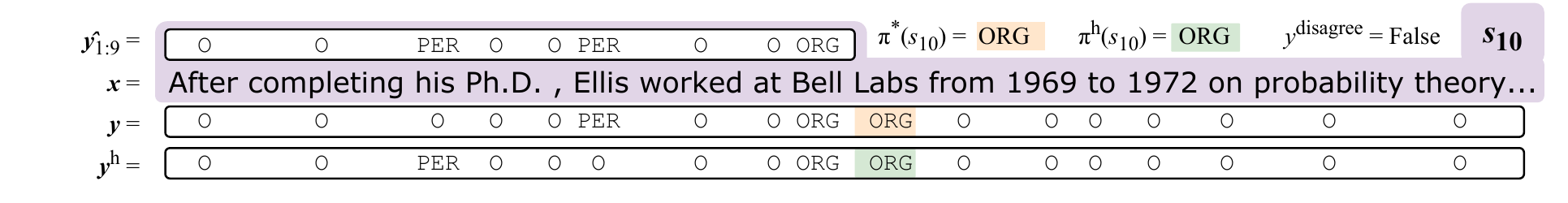}
  \caption{\label{fig:ner}A named entity recognition example (from the Wikipedia page for \href{https://en.wikipedia.org/wiki/Clarence_Ellis_(computer_scientist)}{Clarence Ellis}). $\vx$ is the input sentence and $\vy$ is the (unobserved) ground truth. The predictor $\pi$ operates left-to-right and, in this example, is currently at state $s_{10}$ to tag the $10$th word; the state $s_{10}$  (highlighted in purple) combines $\vx$ with $\hat\vy_{1:9}$. The heuristic makes two errors at $t=4$ and $t=6$. The heuristic label at $t=10$ is $y^h_{10} = $\texttt{ORG}. Under Hamming loss, the cost at $t=10$ is minimized for $a=\texttt{ORG}$, which is therefore the expert action (if it were queried). The label that would be provided for $s_{10}$ to the difference classifier is $0$ because the two policies agree.}
\end{figure*}

We reduce this annotation cost by only asking an expert for labels that are truly needed; 
our algorithm, \ournamelong (\ourname, /\ournameipa/)\footnote{Code is available at: \url{https://github.com/xkianteb/leaqi}} achieves this by capitalizing on two factors.
First, as is typical in active learning (see \autoref{sec:related}), \ourname only asks the expert for a label when it is uncertain.
Second, \ourname assumes access to a \emph{noisy heuristic} labeling function (for instance, a rule-based model, dictionary, or inexpert annotator) that can provide low-quality labels.
\ourname operates by always asking this heuristic for a label, and only querying the expert when it thinks the expert is likely to disagree with this label.
It trains, simultaneously, a \emph{difference classifier} ~\citep{zhang2015active} that predicts disagreements between the expert and the heuristic (see \autoref{fig:ner}).

The challenge in learning the difference classifier is that it must learn based on one-sided feedback: if it predicts that the expert is likely to agree with the heuristic, the expert is not queried and the classifier cannot learn that it was wrong.
We address this one-sided feedback problem using the Apple Tasting framework~\citep{Helmbold2000}, in which errors (in predicting which apples are tasty) are only observed when a query is made (an apple is tasted).
Learning in this way particularly important in the general case where the heuristic is likely not just to have high variance with respect to the expert, but is also statistically biased.

Experimentally (\autoref{sec:results}), we consider three structured prediction settings, each using a different type of heuristic feedback. 
We apply \ourname to: 
English named entity recognition where the heuristic is a rule-based recognizer using gazetteers \citep{cogcompnlp2018}; 
English scientific keyphrase extraction, where the heuristic is an unsupervised method \citep{florescu2017positionrank};
and Greek part-of-speech tagging, where the heuristic is a small dictionary compiled from the training data \citep{zesch2008extracting, Haghighi2006}. 
In all three settings, the expert is a simulated human annotator.
We train \ourname on all three tasks using fixed BERT \citep{devlin2019bert} features, training only the final layer (because we are in the regime of small labeled data).
The goal in all three settings is to minimize the number of words the expert annotator must label.
In all settings, we're able to establish the efficacy of \ourname, showing that it can indeed provide significant label savings over using the expert alone and over several baselines and ablations that establish the importance of both the difference classifier and the Apple Tasting paradigm.

%% file: related.tex
We review first the use of imitation learning for structured prediction, then online active learning, and finally applications of active learning to structured prediction and imitation learning problems.

\subsection{Learning to Search} \label{sec:ilsp}

The learning to search approach to structured prediction casts the joint prediction problem of producing a complex
output as a sequence of smaller classification problems \citep{ratnaparkhi1996maximum,collins04incremental,daume09searn}.
For instance, in the named entity recognition example from \autoref{fig:ner}, an input sentence $\vx$ is labeled one word at a time, left-to-right. 
At the depicted state ($s_{10}$), the model has labeled the first nine words and must next label the tenth word. 
Learning to search approaches assume access to an oracle policy $\pistar$, which provides the optimal label at every position.

In (interactive) imitation learning, we aim to imitate the behavior of the expert policy, $\pistar$, which provides the true labels. The learning to search view allows us to cast structured prediction as a (degenerate) imitation learning task, where states are (input, prefix) pairs,  actions are operations on the output, and the horizon $T$ is the length of the sequence.
States are denoted $s \in \cS$, actions are denoted $a \in [K]$, where $[K] = \{1, \dots, K\}$, and the policy class is denoted $\Pi \subseteq [K]^\cS$.
The goal in learning is to find a policy $\pi \in \Pi$ with small loss on the distribution of states that it, itself, visits. %

A popular imitation learning algorithm, DAgger~\citep{ross11dagger}, is summarized in \autoref{alg:dagger}.
In each iteration, DAgger executes a mixture policy and, at each visited state, queries the expert's action.
This produces a classification example, where the input is the state and the label is the expert's action.
At the end of each iteration, the learned policy is updated by training it on the accumulation of all generated data so far.
DAgger is effective in practice and enjoys appealing theoretical properties; for instance, if the number of iterations $N$ is $\tilde O(T^2 \log (1/\delta))$ then with probability at least $1-\delta$, the generalization error of the learned policy is $O(1/T)$ \citep[Theorem 4.2]{ross11dagger}. 

\definecolor{dimblack}{gray}{0.35}
\begin{algorithm}[t]
  \caption{\label{alg:dagger}$\text{DAgger}(\Pi, N, \langle \be_i \rangle_{i=0}^N, \pistar)$}
  \begin{algorithmic}[1]
    \STATE initialize dataset $D = \{\}$
    \STATE initialize policy $\hat\pi_1$ to any policy in $\Pi$
    \FOR{$i = 1 \dots N$}
      \STATE \textcolor{dimblack}{$\triangleright$ \emph{stochastic mixture policy}}
      \STATE Let $\pi_i = \be_i \pistar + (1-\be_i) \hat\pi_i$ %
      \STATE Generate a $T$-step trajectory using $\pi_i$
      \STATE Accumulate data $D \leftarrow D \cup \{ (s, \pistar(s)) \}$ for all $s$ in those trajectories
      \STATE Train classifier $\hat\pi_{i+1} \in \Pi$ on $D$
    \ENDFOR
    \RETURN best (or random) $\hat\pi_i$
  \end{algorithmic}
\end{algorithm}

\subsection{Active Learning} \label{sec:al}

Active learning has been considered since at least the 1980s often under the name ``selective sampling'' \citep{rendell1986general,atlas1990training}. 
In agnostic online active learning for classification, a learner operates in rounds~\citep[e.g.][]{balcan06agnostic,beygelzimer09active,beygelzimer10agnostic}.
At each round, the learning algorithm is presented an example $\vx$ and must predict a label; the learner must decide whether to query the true label.
An effective margin-based approach for online active learning is provided by \citet{cesa-bianchi06selective}
for linear models. Their algorithm defines a sampling
probability $\rho = b/(b+z)$, where $z$ is the margin on the current example, and $b>0$ is a hyperparameter that controls the aggressiveness of sampling. With probability $\rho$, the algorithm requests the label and performs a perceptron-style update.

Our approach is inspired by \citepos{zhang2015active} setting, where two labelers are available: a free weak labeler and an expensive strong labeler.
Their algorithm minimizes queries to the strong labeler, by learning a difference classifier that predicts, for each example, whether the weak and strong labelers are likely to disagree.
Their algorithm trains this difference classifier using an example-weighting strategy to ensure that its Type II error is kept small, 
establishing statistical consistency, and bounding its sample complexity. 

This type of learning from one-sided feedback falls in the general framework of \emph{partial-monitoring games}, a framework for sequential decision making with imperfect feedback.
Apple Tasting is a type of partial-monitoring game \citep{Littlestone1989WMA}, where, at each round, a learner is presented with an example $\vx$ and must predict a label $\hat y \in \{-1,+1\}$.
After this prediction, the true label is revealed \emph{only} if the learner predicts $+1$.
This framework has been applied in several settings, such as spam filtering and document classification with minority class distributions~\citep{sculley2007}. 
\citet{sculley2007} also conducts a through comparison of two methods that can be used to address the one-side feedback problem: label-efficient online learning \citep{cesa-bianchi06selective} and margin-based learning \citep{vapnik1982}.

\subsection{Active Imitation \& Structured Prediction}

In the context of structured prediction for natural language processing, active learning has been considered both for requesting full structured outputs~\citep[e.g.][]{thompson99parsing,culotta05effort,hachey2005investigating} and for requesting only pieces of outputs~\citep[e.g.][]{Ringger2007ActiveLF,bloodgood2010bucking}.
For sequence labeling tasks, \citet{haertel2008assessing} found that labeling effort depends both on the number of words labeled (which we model), plus a fixed cost for reading (which we do not).

In the context of imitation learning, active approaches have also been considered for at least three decades, often called ``learning with an external critic'' and ``learning by watching'' \citep[][]{whitehead1991study}.
More recently, \citet{judah2012active} describe \emph{RAIL}, an active learning-for-imitation-learning algorithm akin to our \AExp baseline, but which in principle would operate with any underlying i.i.d. active learning algorithm
(not just our specific choice of uncertainty sampling).

%% file: method.tex
Our goal is to learn a structured prediction model with minimal human expert supervision, effectively by combining human annotation with a noisy heuristic.
We present \ourname to achieve this.
As a concrete example, return to \autoref{fig:ner}: at $s_{10}$, $\pi$ must predict the label of the tenth word.
If $\pi$ is confident in its own prediction, \ourname can avoid any query, similar to traditional active learning.
If $\pi$ is not confident, then \ourname considers the label suggested by a noisy heuristic (here: \texttt{ORG}).
\ourname predicts whether the true expert label is likely to disagree with the noisy heuristic.
Here, it predicts no disagreement and avoids querying the expert.

\definecolor{dimblack}{gray}{0.35}

\begin{algorithm}[t]
  \caption{\label{alg:leaqi}$\text{\ourname}(\Pi, \cH, N, \pistar, \piref, b)$}
  \begin{algorithmic}[1]
    \algsetup{linenosize=\tiny}
    \STATE initialize dataset $D = \{\}$
    \STATE initialize policy $\pi_1$ to any policy in $\Pi$
    \STATE initialize difference dataset $S = \{\}$
    \STATE initialize difference classifier $h_1(s) = 1 ~(\forall s)$ \label{l:inith}
    \FOR{$i = 1 \dots N$}
      \STATE Receive input sentence $\vx$
      \STATE \textcolor{dimblack}{$\triangleright$ \emph{generate a $T$-step trajectory using $\pi_i$}} \label{l:rollin}
      \STATE Generate output $\hat\vy$ using $\pi_i$
      \FOR{each $s$ in $\hat\vy$}
        \STATE \textcolor{dimblack}{$\triangleright$ \emph{draw bernouilli random variable}}
        \STATE $Z_{i} \sim {\rm Bern}\left(\frac b {b + \textrm{certainty}(\pi_i,s)}\right )$; see \autoref{sec:certainty} \label{line:prob}
        \IF{ $Z_{i}$ = 1}\label{line:if} \label{l:uncertain}
          \STATE \textcolor{dimblack}{$\triangleright$ \emph{set difference classifier prediction}}
          \STATE $\hat d_{i} = h_i(s)$
          \IF{AppleTaste($s$, $\piref(s)$, $\hat d_{i}$)}\label{l:appletaste}
              \STATE \textcolor{dimblack}{$\triangleright$ \emph{predict agree query heuristic}} 
              \STATE $D \leftarrow D \cup \left\{ ~\left(s, \piref(s)\right)~ \right\}$ \label{l:agree}
          \ELSE
              \STATE \textcolor{dimblack}{$\triangleright$ \emph{predict disagree query expert}}
              \STATE $D \leftarrow D \cup \left\{ ~\left(s, \pistar(s)\right)~ \right\}$ \label{l:disagree}
              \STATE $d_i = \mathbbm{1}\big[\pistar(s) = \piref(s)\big]$
              \STATE $S \leftarrow S \cup \big\{ ~\big(s, \piref(s), \hat d_i, d_i \big)~ \big\}$ %
          \ENDIF
        \ENDIF \label{line:ifend}
      \ENDFOR
      \STATE Train policy $\pi_{i+1} \in \Pi$ on $D$ \label{l:trainpi}
      \STATE Train difference classifier $h_{i+1} \in \cH$ on $S$ to minimize Type II errors (see \autoref{sec:apple}) \label{line:diff}
    \ENDFOR
    \RETURN best (or random) $\pi_i$
  \end{algorithmic}
\end{algorithm}

\subsection{\ournamelong} \label{sec:algorithm}

Our algorithm, \ourname, is specified in \autoref{alg:leaqi}. 
As input, \ourname takes a policy class $\Pi$, a hypothesis class $\cH$ for the difference classifier (assumed to be symmetric and to contain the ``constant one'' function), a number of episodes $N$, an expert policy $\pistar$, a heuristic policy $\piref$, and a confidence parameter $b>0$.
The general structure of \ourname follows that of DAgger, but with three key differences:

\begin{enumerate}[label=(\alph*)]
\item roll-in (\autoref{l:rollin}) is according to the learned policy (not mixed with the expert, as that would require additional expert queries),
\item actions are queried only if the current policy is uncertain at $s$ (\autoref{l:uncertain}), and
\item the expert $\pistar$ is only queried if it is predicted to disagree with the heuristic $\piref$ at $s$ by the difference classifier, or if apple tasting method switches the difference classifier label (\autoref{l:appletaste}; see \autoref{sec:apple}). 
\end{enumerate}
In particular, at each state visited by $\pi_i$, \ourname estimates $z$, the certainty of $\pi_i$'s prediction at that state (see \autoref{sec:certainty}).
A sampling probability $\rho$ is set to $b/(b+z)$ where $z$ is the certainty, and so if the model is very uncertain then $\rho$ tends to zero, following  \citep{cesa-bianchi06selective}.
With probability $\rho$, \ourname will collect \emph{some} label.

When a label is collected (\autoref{l:uncertain}), the difference classifier $h_i$ is queried on state $s$ to predict if $\pistar$ and $\piref$ are likely to disagree on the correct action. (Recall that $h_1$ always predicts disagreement per \autoref{l:inith}.)
The difference classifier's prediction, $\hat d_i$, is passed to an \emph{apple tasting} method in \autoref{l:appletaste}.
Intuitively, most apple tasting procedures (including the one we use, STAP; see \autoref{sec:apple}) return $\hat d_i$, unless the difference classifier is making many Type II errors, in which case it may return $\lnot \hat d_i$.

\newcommand{\aref}{a^{\text{h}}}

\begin{algorithm}[t]
  \caption{\label{alg:AppleTaste:STAP}$\text{AppleTaste\_STAP}(S, \aref_i, \hat d_i)$}
  \begin{algorithmic}[1]
    \STATE \textcolor{dimblack}{$\triangleright$ \emph{count examples that are action $\aref_i$}}
    \STATE let $t = \sum_{(\_,a,\_,\_) \in S} \mathbbm{1}[\aref_i = a]$ %
    \STATE \textcolor{dimblack}{$\triangleright$ \emph{count mistakes made on action $\aref_i$}}
    \STATE let $m = \sum_{(\_,a,\hat d,d) \in S} \mathbbm{1}[\hat d \neq d \land \aref_i = a]$%
    \STATE $w = $ $\frac{t}{|S|}$ 
    \quad\textcolor{dimblack}{$\triangleright$ \emph{percentage of time $\aref_i$ was seen}}
    \IF{$w$ < 1}
        \STATE \textcolor{dimblack}{$\triangleright$ \emph{skew distribution}}
        \STATE draw $r \sim \textit{Beta}(1-w,1)$
    \ELSE
        \STATE draw $r \sim \textit{Unifomm}(0,1)$
    \ENDIF
    \RETURN 
      $(d=1) \land (r \leq \sqrt{(m +1)/t})$ \label{line:appleradom}
  \end{algorithmic}
\end{algorithm}

A target action is set to $\piref(s)$ if the apple tasting algorithm returns ``agree'' (\autoref{l:agree}), and the expert $\pistar$ is only queried if disagreement is predicted (\autoref{l:disagree}).
The state and target action (either heuristic or expert) are then added to the training data.
Finally, if the expert was queried, then a new item is added to the difference dataset, consisting of the state, the heuristic action on that state, the difference classifier's prediction, and the ground truth for the difference classifier whose input is $s$ and whose label is whether the expert and heuristic \emph{actually} disagree.
Finally, $\pi_{i+1}$ is trained on the accumulated action data, and $h_{i+1}$ is trained on the difference dataset (details in \autoref{sec:certainty}).

There are several things to note about \ourname:
\begin{enumerate}[label=$\diamond$]
\item If the current policy is already very certain, a expert annotator is \emph{never} queried.
\item If a label is queried, the expert is queried only if the difference classifier predicts disagreement with the heuristic, or the apple tasting procedure flips the difference classifier prediction.
\item Due to apple tasting, most errors the difference classifier makes will cause it to query the expert unnecessarily; this is the ``safe'' type of error (increasing sample complexity but not harming accuracy), versus a Type II error (which leads to biased labels).
\item The difference classifier is only trained on states where the policy is uncertain, which is exactly the distribution on which it is run.
\end{enumerate}

\subsection{Apple Tasting for One-Sided Learning}\label{sec:apple}
The difference classifier $h \in \cH$ must be trained (line~\ref{line:diff}) based on one-sided feedback (it only observes errors when it predicts ``disagree``) to minimize Type II errors (it should only very rarely predict ``agree'' when the truth is ``disagree''). 
This helps keep the labeled data for the learned policies unbiased. 
The main challenge here is that the feedback to the difference classifier is \emph{one-sided}: that is, if it predicts ``disagree'' then it gets to see the truth, but if it predicts ``agree'' it never finds out if it was wrong. 
We use one of \citep{Helmbold2000}'s algorithms, STAP (see \autoref{alg:AppleTaste:STAP}), which works by random sampling from apples that are predicted to not be tasted and tasting them anyway (line~\ref{line:appleradom}). 
Formally, STAP tastes apples that are predicted to be bad with probability $\sqrt{(m +1)/t}$, where $m$ is the number of mistakes, and $t$ is the number of apples tasted so far.

We adapt Apple Tasting algorithm STAP to our setting for controlling the number of Type II errors made by the difference classifier as follows.
First, because some heuristic actions are much more common than others, we run a separate apple tasting scheme \emph{per} heuristic action (in the sense that we count the number of error \emph{on this heuristic action} rather than globally).
Second, when there is significant action imbalance\footnote{For instance, in named entity recognition, both the heuristic and expert policies label the majority of words as \texttt{O} (not an entity). 
As a result, when the heuristic says \texttt{O}, it is very likely that the expert will agree. 
However, if we aim to optimize for something other than accuracy---like F1---it is precisely these disagreements that we need to find.} we find it necessary to skew the distribution from STAP more in favor of querying.
We achieve this by sampling from a \emph{Beta} distribution (generalizing the uniform), whose mean is shifted toward zero for more frequent heuristic actions. 
This increases the chance that Apple Tasting will have on finding bad apples error for each action (thereby keeping the false positive rate low for predicting disagreement).

\subsection{Measuring Policy Certainty} \label{sec:certainty}
In step~\ref{line:prob}, \ourname must estimate the certainty of $\pi_i$ on $s$. Following~\citet{cesa-bianchi06selective}, we implement this using a margin-based criteria. To achieve this, we consider $\pi$ as a function that maps actions to scores and then chooses the action with largest score. The certainty measure is then the difference in scores between the highest and second highest scoring actions:
\begin{align*} \label{eq:certainty}
  \text{certainty}(\pi, s) &= \max_a \pi(s,a) - \max_{a' \neq a} \pi(s,a')
\end{align*}

\subsection{Analysis} \label{sec:analysis}

Theoretically, the main result for \ourname is an interpretation of the main DAgger result(s).
Formally, let $d_\pi$ denote the distribution of states visited by $\pi$, $C(s,a) \in [0,1]$ be the immediate cost of performing action $a$ in state $s$, $C_\pi(s) = \Ep_{a\sim \pi(s)} C(s,a)$, and the total expected cost of $\pi$ to be $J(\pi) = T \Ep_{s\sim d_\pi} C_\pi(s)$, where $T$ is the length of trajectories. $C$ is not available to a learner in an imitation setting; instead the algorithm observes an expert and minimizes a surrogate loss $\ell(s,\pi)$ (e.g., $\ell$ may be zero/one loss between $\pi$ and $\pistar$). We assume $\ell$ is strongly convex and bounded in $[0,1]$ over $\Pi$.

Given this setup assumptions, let $\epspol = \min_{\pi \in \Pi} \avgi \Epi \ell(s, \pi)$ be the true loss of the best policy in hindsight, let $\epsdc = \min_{h \in \cH} \avgi \Epi \err(s, h, \pistar(s)\neq \piref(s))$ be the true error of the best difference classifier in hindsight, and assuming that the regret of the policy learner is bounded by $\regpol(N)$ after $N$ steps, \citet{ross11dagger} shows the following\footnote{Proving a stronger result is challenging: analyzing the sample complexity of an active learning algorithm that uses a difference classifier---even in the non-sequential setting---is quite involved~\citep{zhang2015active}.}:

\begin{theorem}[Thm 4.3 of \citet{ross11dagger}]
  After $N$ episodes each of length $T$, under the assumptions above, with probability at least $1-\delta$ there exists a policy $\pi \in \pi_{1:N}$ such that:
  \begin{align*}
    &\Ep_{s \sim d_\pi} \ell(s, \pi) \leq \\
     & \quad \epspol + \regpol(N) + \sqrt{(2/N)\log (1/\delta)}
  \end{align*}
\end{theorem}

\begin{table*}[t]
  \newcolumntype{L}[1]{>{\raggedright\arraybackslash}p{#1}}
  \centering
  \rowcolors{2}{gray!15}{white}
  \begin{tabular}{lL{1.72in}L{1.59in}L{1.59in}}
    \toprule
    \textbf{Task}
    & Named Entity Recognition
    & Keyphrase Extraction 
    & Part of Speech Tagging 
    \\
    \midrule
    \textbf{Language}
    & English (en)
    & English (en)
    & Modern Greek (el)
    \\
    \textbf{Dataset}
    & CoNLL'03 \citep{TjongKimSang2003}
    & SemEval 2017 Task 10 \citep{augenstein2017semeval}
    & Universal Dependencies \cite{Nivre2018UD}
    \\
    \textbf{\# Ex}
    & $14,987$
    & $2,809$
    & $1,662$
    \\
    \textbf{Avg. Len}
    & $14.5$
    & $26.3$
    & $25.5$
    \\
    \textbf{\# Actions}
    & $5$
    & $2$
    & $17$
    \\
    \textbf{Metric}
    & Entity F-score
    & Keyphrase F-score
    & Per-tag accuracy
    \\
    \textbf{Features}
    & English BERT~\citep{devlin2019bert}
    & SciBERT~\citep{Beltagy2019SciBERT}
    & M-BERT~\citep{devlin2019bert}\\
    \textbf{Heuristic}
    & String matching against an offline gazeteer of entities from \citet{cogcompnlp2018}
    & Output from an unsupervised keyphrase extraction model \citet{florescu2017positionrank}
    & Dictionary from Wiktionary, similar to \citet{zesch2008extracting} and \citet{Haghighi2006}
    \\
    \textbf{Heur Quality}
    & P $88\%$, R $27\%$, F $41\%$
    & P $20\%$, R $44\%$, F $27\%$
    & $10\%$ coverage, $67\%$ acc \\
    \bottomrule
  \end{tabular}
  \caption{\label{tab:tasks}An overview of the three tasks considered in experiments.}
\end{table*}

\noindent
This holds regardless of how $\pi_{1:N}$ are trained (\autoref{l:trainpi}). The question of how well \ourname performs becomes a question of how well the combination of uncertainty-based sampling and the difference classifier learn. So long as those do a good job on their individual \emph{classification} tasks, DAgger guarantees that the \emph{policy} will do a good job. This is formalized below, where $Q^\star(s,a)$ is the best possible cumulative cost (measured by $C$) starting in state $s$ and taking action $a$:

\begin{theorem}[Theorem 2.2 of \citet{ross11dagger}]
  Let $u$ be such that $Q^\star(s,a) - Q^\star(s,\pistar(s)) \leq u$ for all $a$ and all $s$ with $d_\pi(s) > 0$; then for some $\pi \in \pi_{1:N}$, as $N \rightarrow \infty$:
  \begin{align*}
    J(\pi) &\leq J(\pistar) + u T \epspol
  \end{align*}
\end{theorem}

\noindent
Here, $u$ captures the most long-term impact a single decision can have; for example, for average Hamming loss, it is straightforward to see that $u = \frac 1 T$ because any single mistake can increase the number of mistakes by at most $1$. For precision, recall and F-score, $u$ can be as large as one in the (rare) case that a single decision switches from one true positive to no true positives.

%% file: applications.tex
The primary research questions we aim to answer experimentally are:
\begin{enumerate}[label=Q\arabic*\xspace]
  \item Does uncertainty-based active learning achieve lower query complexity than passive learning in the learning to search settings? \label{q:actvspass}
  \item Does learning a difference classifier improve query efficiency over active learning alone? \label{q:diff}
  \item Does Apple Tasting successfully handle the problem of learning from one-sided feedback?\label{q:apple}
  \item Is the approach robust to cases where the noisy heuristic is uncorrelated with the expert? \label{q:robust}
  \item Is casting the heuristic as a policy more effective than using its output as features? \label{q:feat}
\end{enumerate}
To answer these questions, we conduct experiments on three tasks (see \autoref{tab:tasks}):
English named entity recognition, English scientific keyphrase extraction, and low-resource part of
speech tagging on Modern Greek (el), selected as a low-resource setting.

\subsection{Algorithms and Baselines}
\label{sec:baselines}

In order to address the research questions above, we compare \ourname to several baselines. The baselines below compare our approach to previous methods:

\begin{description}
  \item[\PExp.] Passive DAgger (\autoref{alg:dagger})
  \item[\AExp.] An active variant of DAgger that asks for labels only when uncertain. (This is equivalent to \ourname, but with neither the difference classifier nor apple tasting.)
  \item[\PRef] \PExp with the heuristic policy's output appended as an input feature.
  \item[\ARef] \AExp with the heuristic policy as a feature.
\end{description}

\noindent
The next set of comparisons are explicit ablations:

\begin{description}
  \item[\OurNoAt] \ourname with no apple tasting. 
  \item[\RandRef] \ourname, but where the heuristic returns a label uniformly at random.
\end{description}

\noindent
The baselines and \ourname share a linear relationship. 
\PExp is the baseline algorithm used by all algorithms described above but it is very query inefficient with respect to an expert annotator. 
\AExp introduces active learning to make \PExp more query efficient; the delta to the previous addresses \ref{q:actvspass}.
\OurNoAt introduces the difference classifier; the delta addresses \ref{q:diff}.
\ourname adds apple tasting to deal with one-sided learning; the delta addresses \ref{q:apple}.
Finally, \RandRef (vs \ourname) addresses \ref{q:robust} and the \textsc{+Feat} variants address \ref{q:feat}.

\subsection{Data and Representation} \label{sec:data}
For \emph{named entity recognition}, we use training, validation, and test data from CoNLL'03 \citep{TjongKimSang2003}, consisting of IO tags instead of BIO tags (the ``B'' tag is almost never used in this dataset, so we never attempt to predict it) over four entity types: Person, Organization, Location, and Miscellaneous.
For \emph{part of speech tagging}, we use training and test data from modern Greek portion of the Universal Dependencies (UD) treebanks \citep{Nivre2018UD}, consisting of 17 universal tags\footnote{\scriptsize ADJ, ADP, ADV, AUX, CCONJ, DET, INTJ, NOUN, NUM, PART, PRON, PROPN, PUNCT, SCONJ, SYM, VERB, X.}.
For \emph{keyphrase extraction}, we use training, validation, and test data from SemEval 2017 Task 10 \citep{augenstein2017semeval}, consisting of IO tags (we use one ``I'' tag for all three keyphrase types).

In all tasks, we implement both the policy and difference classifier by fine-tuning the last layer of a BERT embedding representation~\citep{devlin2019bert}.
More specifically, for a sentence of length $T$, $w_1, \dots, w_T$, we first compute BERT embeddings for each word, $\vx_1, \dots, \vx_T$ using the appropriate BERT model: English BERT and M-BERT\footnote{Multilingual BERT ~\citep{devlin2019bert}} for named entity and part-of-speech, respectively, and SciBERT~\citep{Beltagy2019SciBERT} for keyphrase extraction.
We then represent the state at position $t$ by concatenating the word embedding at that position with a one-hot representation of the previous action: $s_t = [ \vw_t ; \text{onehot}(a_{t-1}) ]$.
This feature representation is used both for learning the labeling policy and also learning the difference classifier.

\begin{figure*}[t]
  \minipage{0.32\textwidth}
    \includegraphics[width=\linewidth,height=0.9\textwidth]{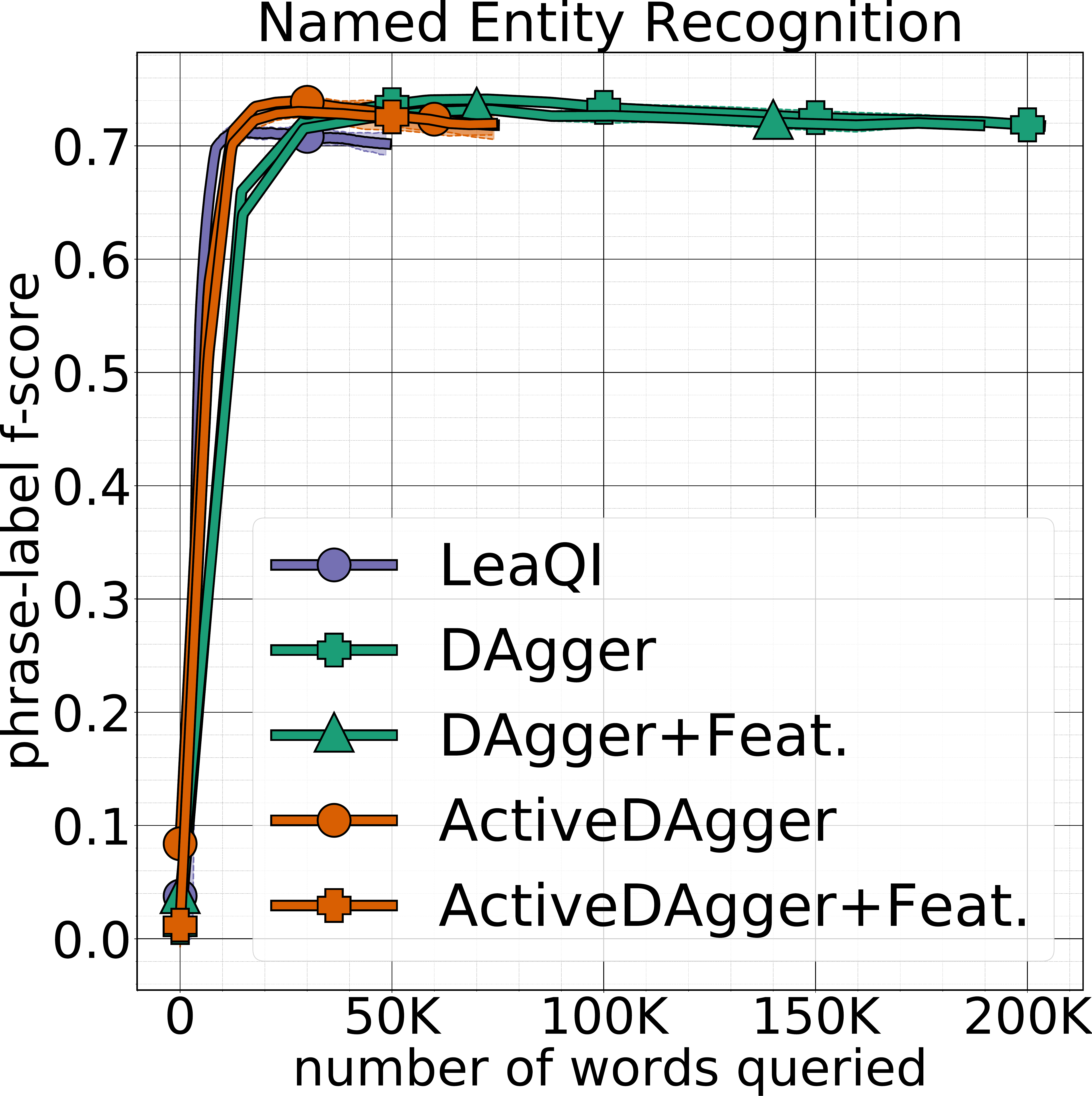}
  \endminipage\hfill
  \minipage{0.32\textwidth}
    \includegraphics[width=\linewidth,height=0.9\textwidth]{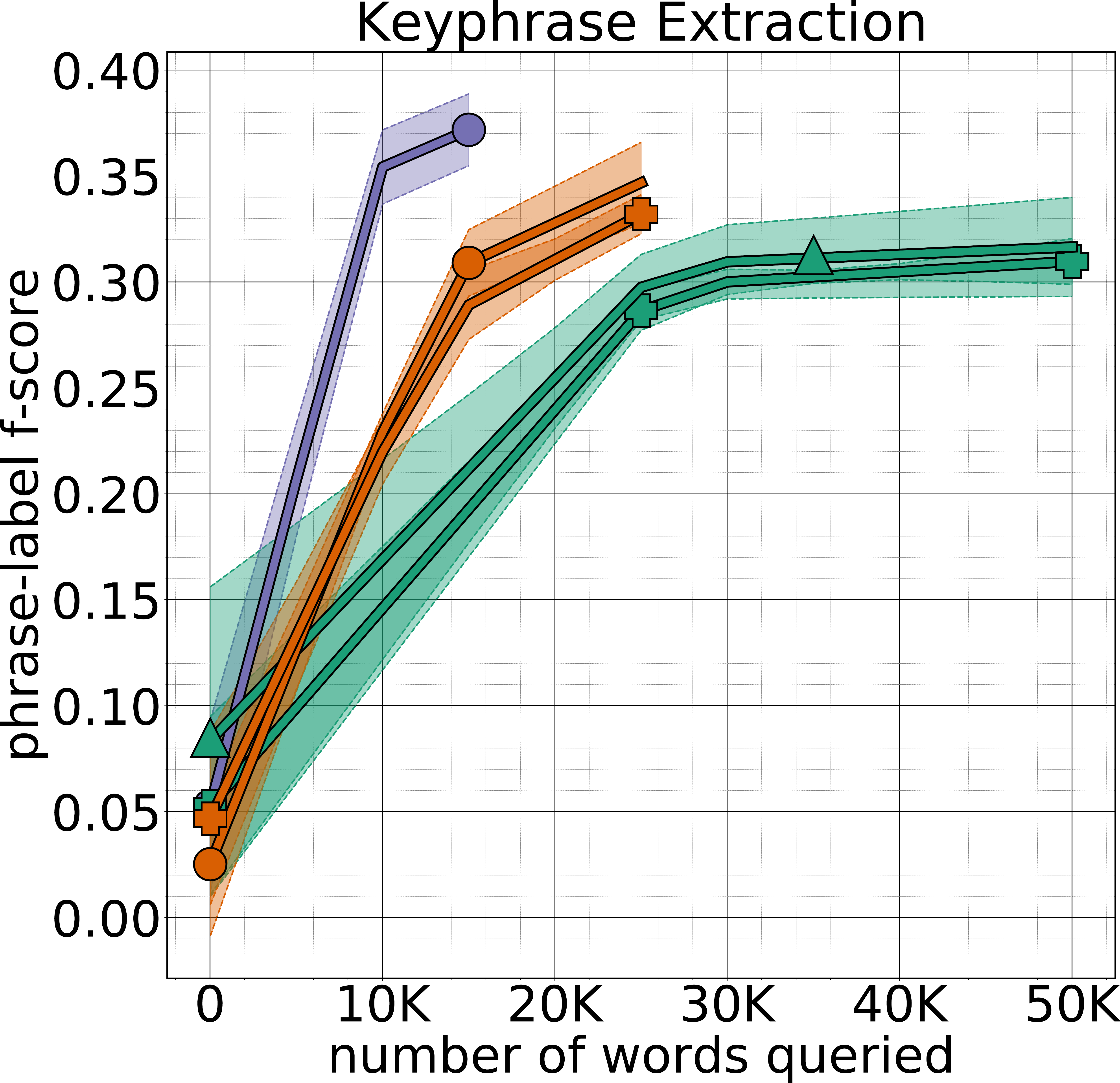}
  \endminipage\hfill
  \minipage{0.32\textwidth}%
    \includegraphics[width=\linewidth,height=0.9\textwidth]{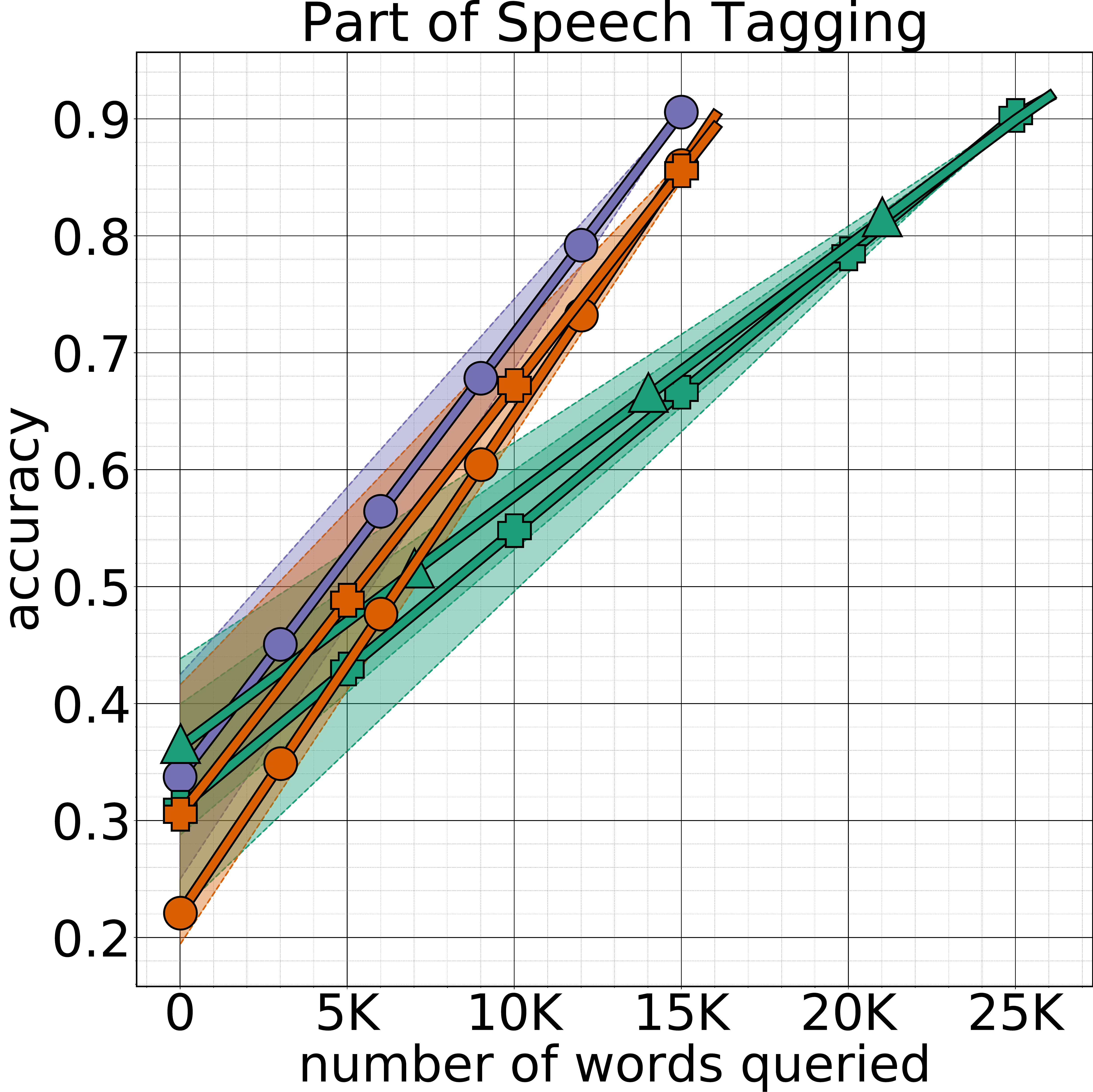}
  \endminipage

  \minipage{0.32\textwidth}
    \includegraphics[width=\linewidth,height=0.9\textwidth]{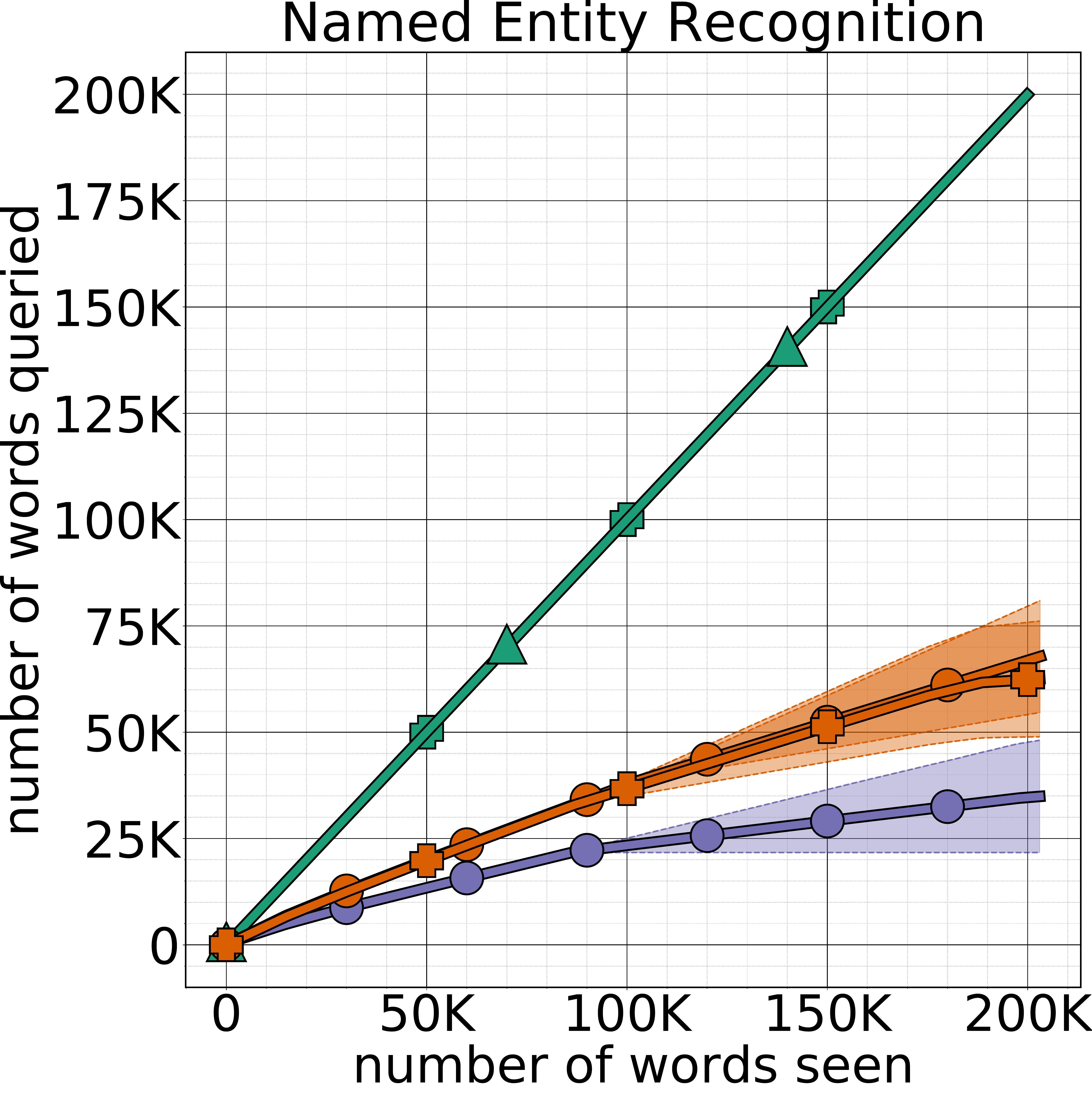}
  \endminipage\hfill
  \minipage{0.32\textwidth}
    \includegraphics[width=\linewidth,height=0.9\textwidth]{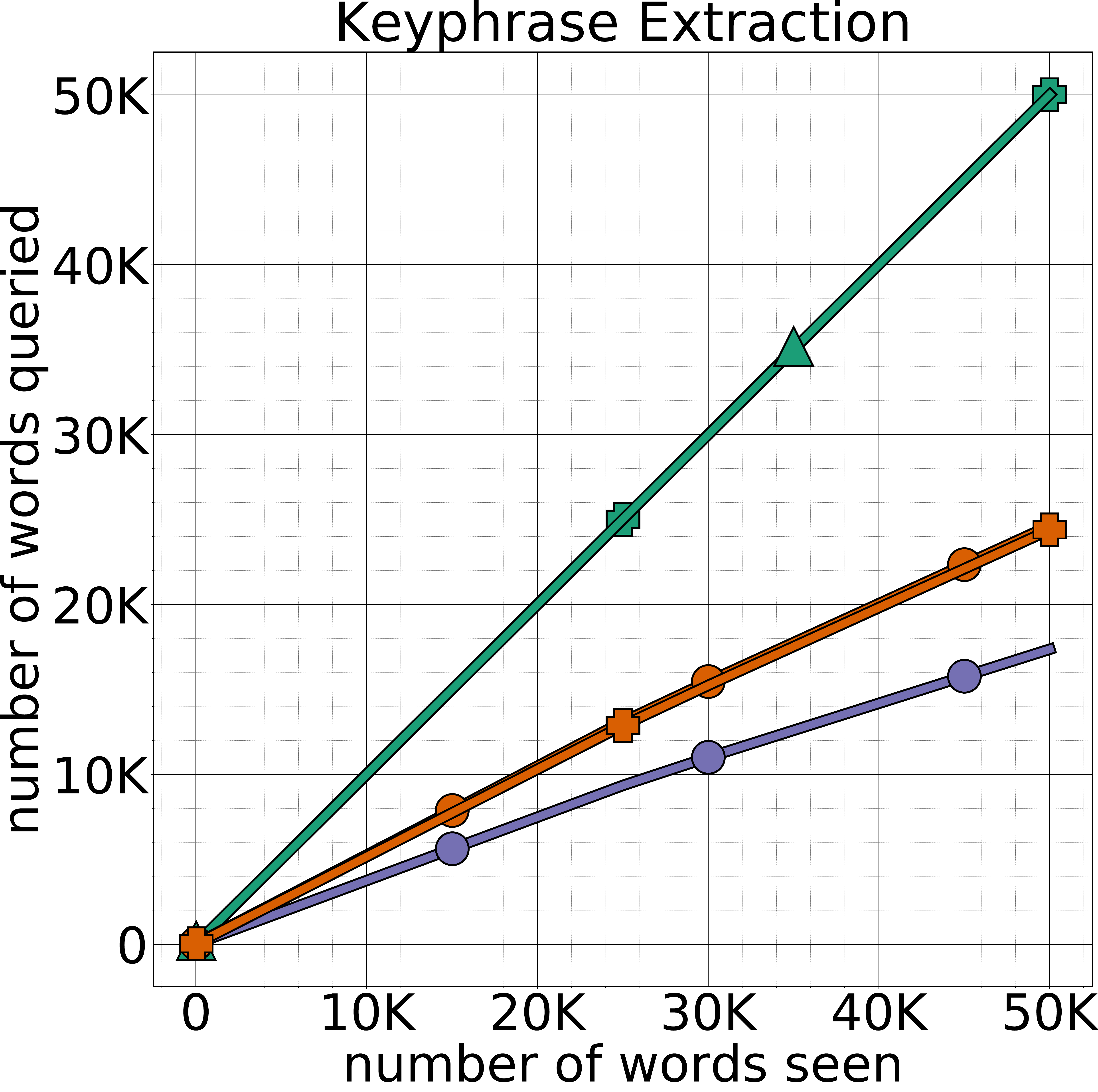}
  \endminipage\hfill
  \minipage{0.32\textwidth}%
    \includegraphics[width=\linewidth,height=0.9\textwidth]{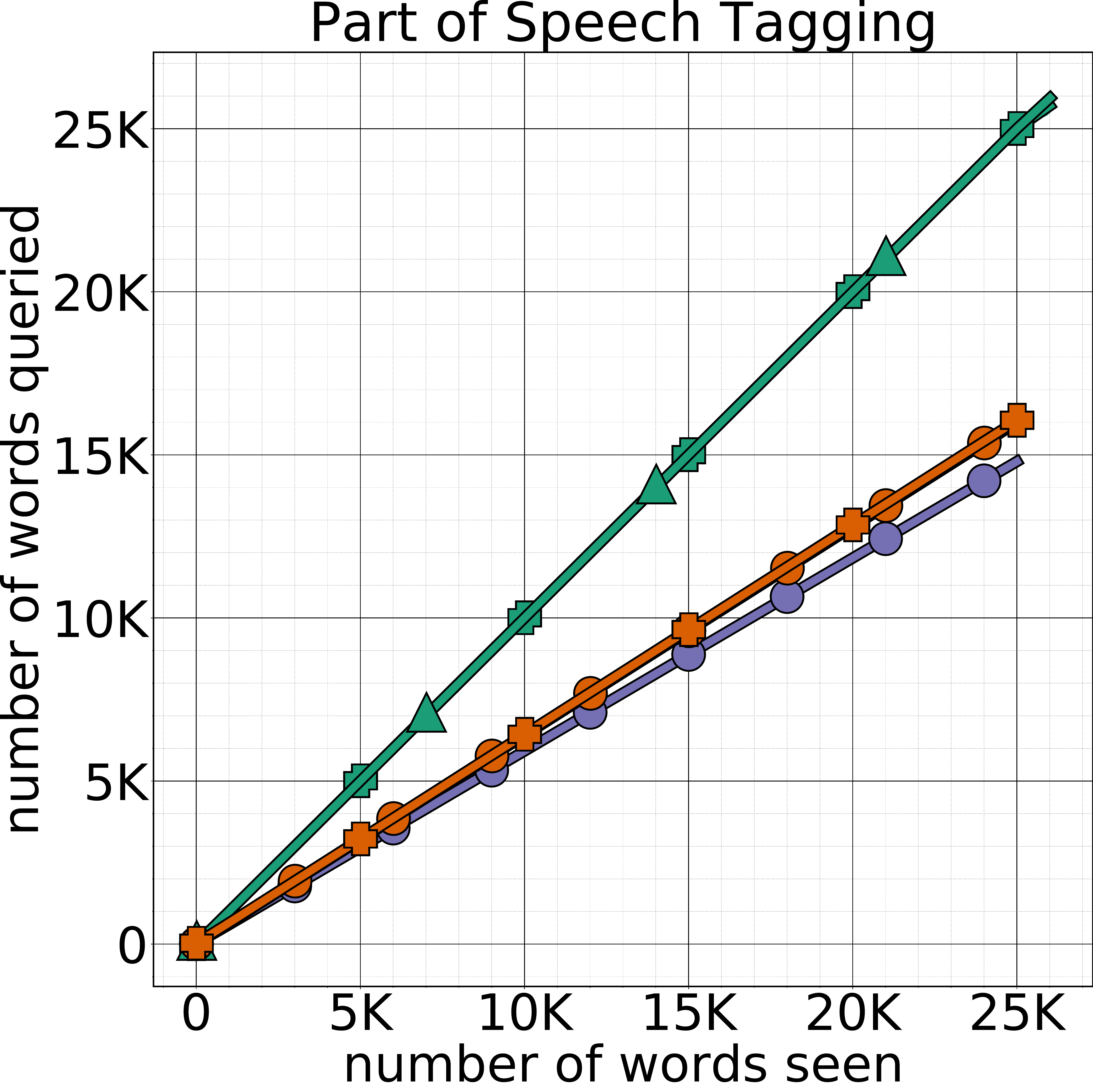}
  \endminipage

  \caption{\label{fig:main} Empirical evaluation on three tasks: (left) named entity recognition, (middle) keyphrase extraction and (right) part of speech tagging. The top rows shows performance (f-score or accuracy) with respect to the number of queries to the expert. The bottom row shows the number of queries as a function of the number of words seen.}
  \end{figure*}

  \begin{figure*}[t]
  \minipage{0.32\textwidth}
    \includegraphics[width=\linewidth,height=0.9\textwidth]{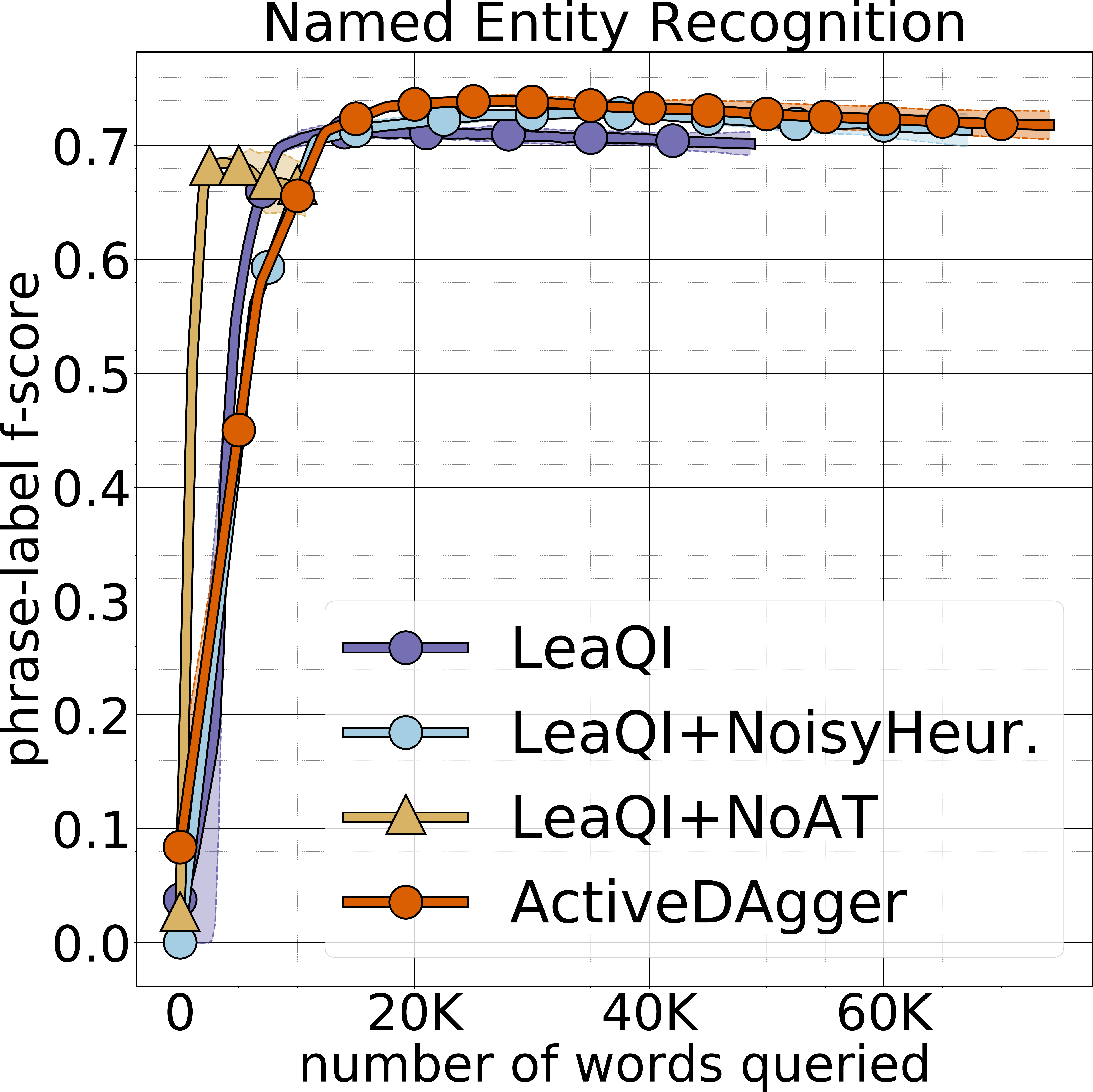}
  \endminipage\hfill
  \minipage{0.32\textwidth}
    \includegraphics[width=\linewidth,height=0.9\textwidth]{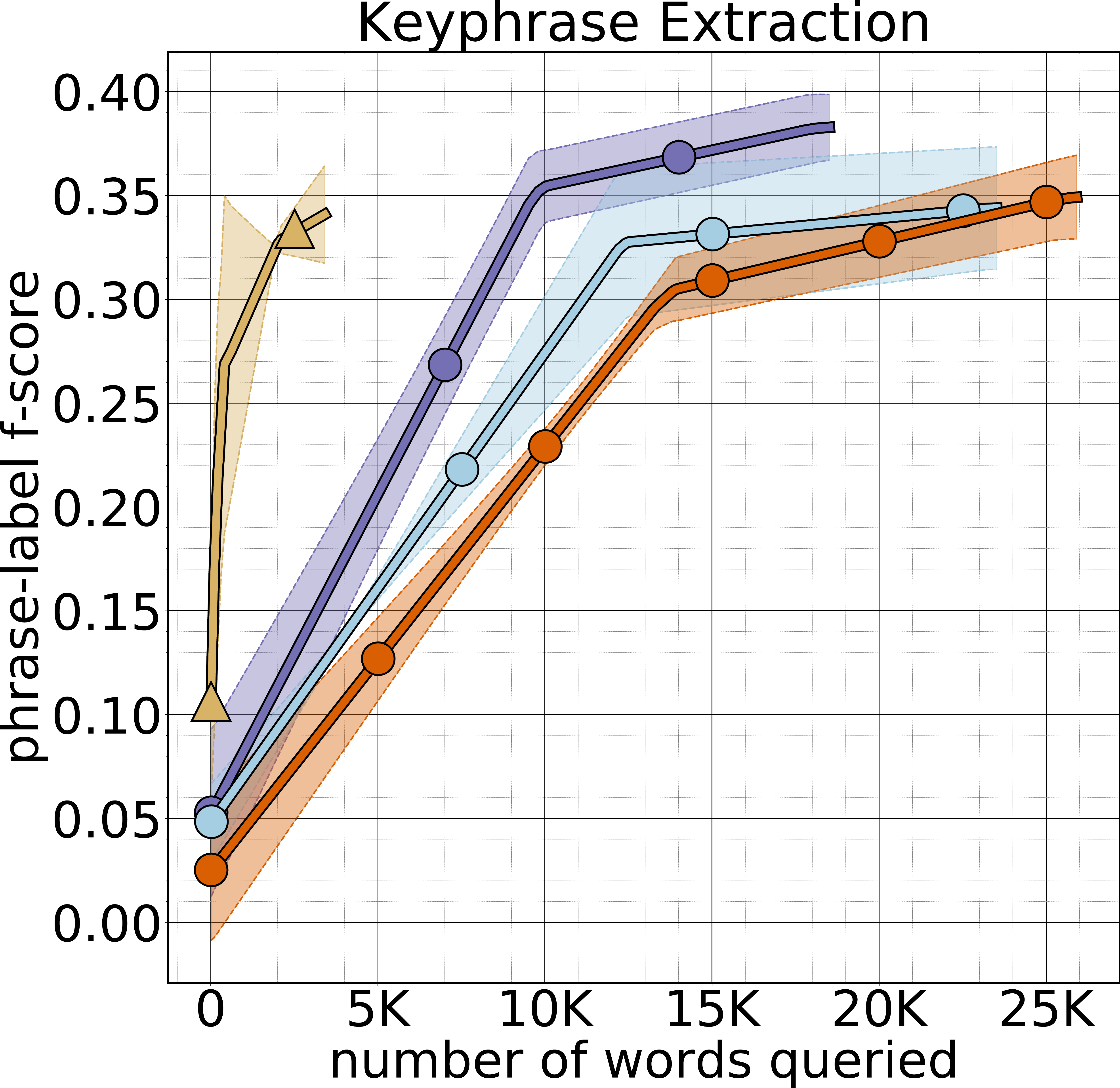}
  \endminipage\hfill
  \minipage{0.32\textwidth}%
    \includegraphics[width=\linewidth,height=0.9\textwidth]{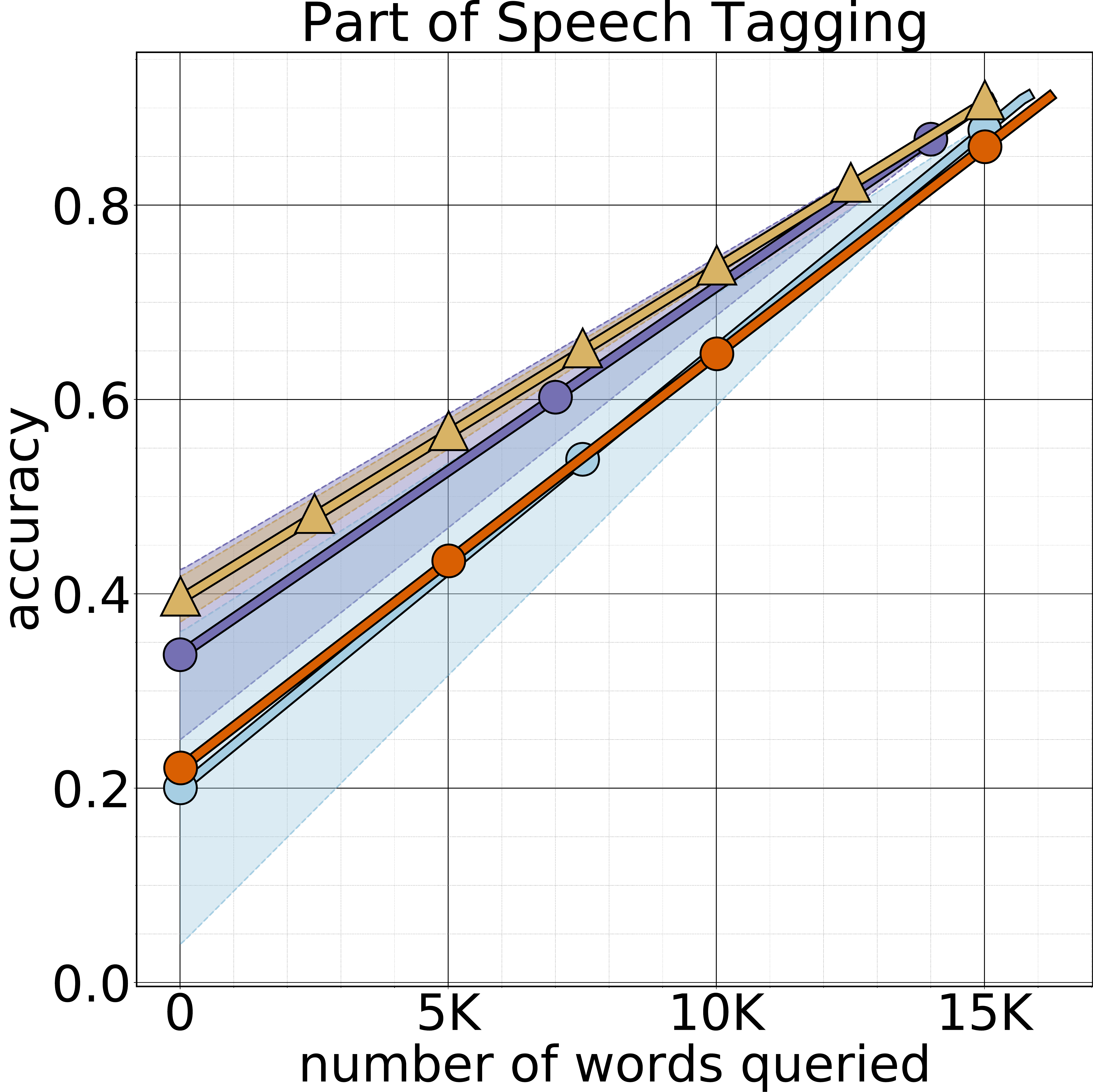}
  \endminipage
  \caption{\label{fig:ablation} Ablation results on (left) named entity recognition, (middle) keyphrase extraction and (right) part of speech tagging. In addition to \ourname and DAgger (copied from \autoref{fig:main}), these graphs also show \OurNoAt (apple tasting disabled), and \RandRef (a heuristic that produces labels uniformly at random).}
  \end{figure*}

\subsection{Expert Policy and Heuristics} \label{sec:ner:policies}

In all experiments, the expert $\pistar$ is a simulated human annotator who annotates one word at a time.
The expert returns the optimal action for the relevant evaluation metric (F-score for named entity recognition and keyphrase extraction, and accuracy for part-of-speech tagging).
We take the annotation cost to be the total number of words labeled.

The heuristic we implement for named entity recognition is a high-precision gazeteer-based string matching approach.
We construct this by taking a gazeteer from Wikipedia using the CogComp framework \citep{cogcompnlp2018}, and use FlashText \citep{Singh2017} to label the dataset.
This heuristic achieves a precision of $0.88$, recall of $0.27$ and F-score of $0.41$ on the training data.

The keyphrase extraction heuristic is the output of an ``unsupervised keyphrase extraction'' approach~\citep{florescu2017positionrank}. This system is a graph-based approach that constructs word-level graphs incorporating positions of all word occurrences information; then using PageRank to score the words and phrases. This heuristic achieves a precision of $0.20$, recall of $0.44$ and F-score of $0.27$ on the training data. 

The part of speech tagging heuristic is based on a small dictionary compiled from Wiktionary.
Following \citet{Haghighi2006} and \citet{zesch2008extracting}, we extract this dictionary using Wiktionary as follows: for word $w$ in our training data, we find the part-of-speech $y$ by querying Wiktionary. If $w$ is in Wikitionary, we convert the Wikitionary part of speech tag to a Universal Dependencies tag (see \autoref{sec:wiktionary}), and if word $w$ is not in Wiktionary, we use a default label of ``X''. Furthermore, if word $w$ has multiple parts of speech, we select the first part of speech tag in the list. The label ``X'' is chosen $90\%$ of the time. For the remaining $10\%$, the heuristic achieves an accuracy of $0.67$ on the training data.

\subsection{Experimental Setup} \label{sec:ner:setup}

Our experimental setup is online active learning.
We make a single pass over a dataset, and the goal is to achieve an accurate system as quickly as possible.
We measure performance (accuracy or F-score) after every $1000$ words ($\approx 50$ sentences) on heldout test data, and produce error bars by averaging across three runs and reporting standard deviations.

Hyperparameters for \PExp are optimized using grid-search on the named entity recognition training data and evaluated on development data. We then fix \PExp hyperparameters for all other experiments and models. The difference classifier hyperparameters are subsequently optimized in the same manner. We fix the difference classifier hyperparameters for all other experiments.\footnote{We note that this is a somewhat optimistic hyperparameter setting: in the real world, model selection for active learning is extremely challenging.
Details on hyperparameter selection and \ourname's robustness across a rather wide range of choices are presented in \autoref{sec:hyperparameters}, \autoref{sec:lr} and \autoref{sec:b} for keyphrase extraction and part of speech tagging.}

\subsection{Experimental Results} 
\label{sec:results}

The main results are shown in the top two rows of \autoref{fig:main}; ablations of \ourname are shown in \autoref{fig:ablation}. In \autoref{fig:main}, the top row shows traditional learning curves (performance vs number of queries), and the bottom row shows the number of queries made to the expert as a function of the total number of words seen.

\paragraph{Active vs Passive (\ref{q:actvspass}).}
In all cases, we see that the active strategies improve on the passive strategies; this difference is largest in keyphrase extraction, middling for part of speech tagging, and small for NER.
While not surprising given previous successes of active learning, this confirms that it is also a useful approach in our setting. As expected, the active algorithms query far less than the passive approaches, and \ourname queries the least.

\paragraph{Heuristic as Features vs Policy (\ref{q:feat}).} We see that while adding the heuristic's output as a feature can be modestly useful, it is not uniformly useful and, at least for keyphrase extraction and part of speech tagging, it is not as effective as \ourname. For named entity recognition, it is not effective at all, but this is also a case where all algorithms perform essentially the same. Indeed, here, \ourname learns quickly with few queries, but never quite reaches the performance of ActiveDAgger. This is likely due to the difference classifier becoming overly confident too quickly, especially on the ``O'' label, given the (relatively well known) oddness in mismatch between development data and test data on this dataset.

\paragraph{Difference Classifier Efficacy (\ref{q:diff}).}
Turning to the ablations (\autoref{fig:ablation}), we can address \ref{q:diff} by comparing the ActiveDAgger curve to the LeaQI+NoAT curve. Here, we see that on NER and keyphrase extraction, adding the difference classifier without adding apple tasting results in a far worse model: it learns very quickly but plateaus much lower than the best results. The exception is part of speech tagging, where apple tasting does not seem necessary (but also does not hurt). Overall, this essentially shows that without controlling Type II errors, the difference classifier on it's own does not fulfill its goals.

\paragraph{Apple Tasting Efficacy (\ref{q:apple}).}
Also considering the ablation study, we can compare LeaQI+NoAT with LeaQI.
In the case of part of speech tagging, there is little difference: using apple tasting to combat issues of learning from one sided feedback neither helps nor hurts performance.
However, for both named entity recognition and keyphrase extraction, removing apple tasting leads to faster learning, but substantially lower final performance (accuracy or f-score).
This is somewhat expected: without apple tasting, the training data that the policy sees is likely to be highly biased, and so it gets stuck in a low accuracy regime.

\paragraph{Robustness to Poor Heuristic (\ref{q:robust}).}
We compare LeaQI+NoisyHeur to ActiveDAgger. Because the heuristic here is useless, the main hope is that it does not \emph{degrade} performance below ActiveDAgger. Indeed, that is what we see in all three cases: the difference classifier is able to learn quite quickly to essentially ignore the heuristic and only rely on the expert.

%% file: discussion.tex
In this paper, we considered the problem of reducing the number of queries to an expert labeler for structured prediction problems.
We took an imitation learning approach and developed an algorithm, \ourname, which leverages a source that has low-quality labels: a heuristic policy that is suboptimal but free.
To use this heuristic as a policy, we learn a difference classifier that effectively tells \ourname when it is safe to treat the heuristic's action as if it were optimal.
We showed empirically---across Named Entity Recognition, Keyphrase Extraction and Part of Speech Tagging tasks---that the active learning approach improves significantly on passive learning, and that leveraging a difference classifier improves on that.
\begin{enumerate}
\item In some settings, learning a difference classifier may be as hard or harder than learning the structured predictor; for instance if the task is binary sequence labeling (e.g., word segmentation), minimizing its usefulness.
\item The true labeling cost is likely more complicated than simply the number of individual actions queried to the expert.
\end{enumerate}

Despite these limitations, we hope that \ourname provides a useful (and relatively simple) bridge that can enable using rule-based systems, heuristics, and unsupervised models as building blocks for more complex supervised learning systems.
This is particularly attractive in settings where we have very strong rule-based systems, ones which often outperform the best statistical systems, like coreference resolution \citep{stanford-coref}, information extraction \citep{riloff2003learning}, and morphological segmentation and analysis \citep{smit2014morfessor}.

%% file: appendix_experiments.tex
\subsection{Wiktionary to Universal Dependencies} \label{sec:wiktionary}

\begin{table*}[!htbp]
  \centering
  \small
  \rowcolors{2}{gray!15}{white}
  \begin{tabular}{lp{1.69in}p{1.65in}p{1.65in}}
    \toprule
    \textbf{POS Tag Source}
    & Greek, Modern (el) Wiktionary
    & Universal Dependencies
    \\
    \midrule
    & adjective
    & ADJ
    \\
    & adposition
    & ADP
    \\
    & preposition
    & ADP
    \\
    & adverb
    & ADV
    \\
    & auxiliary
    & AU
    \\
    & coordinating conjunction
    & CCONJ
    \\
    & determiner
    & DET
    \\
    & interjection
    & INTJ
    \\
    & noun
    & NOUN
    \\
    & numeral
    & NUM 
    \\
    & particle
    & PART
    \\
    & pronoun
    & PRON
    \\
    & proper noun
    & pROPN
    \\
    & punctuation
    & PUNCT
    \\
    & subordinating conjunction
    & SCONJ
    \\
    & symbol
    & SYM
    \\
    & verb
    & VERB
    \\
    & other
    & X
    \\
    & article
    & DET
    \\
    & conjunction
    & PART
    \\
    \bottomrule
  \end{tabular}
  \caption{\label{tab:convert}Conversion between Greek, Modern (el) Wiktionary POS tags and Universal Dependencies POS tags.}\label{pos:converter}
\end{table*}

\subsection{Hyperparameters} \label{sec:hyperparameters}
Here we provide a table of all of hyperparameters we considered for \ourname and baselines models. (see section~\ref{sec:ner:setup})
\begin{table*}[h!]
  \caption{Hyperparameters}
  \centering
  \begin{tabular}{lll}
    Hyperparameter & Values Considered & Final Value \\
    \hline
    Policy Learning rate & $10^{-3}, 10^{-4}, 10^{-5}, 10^{-6}, 5.5\cdot 10^{-6}, 10^{-6}$ & $10^{-6}$\\
    Difference Classifier Learning rate $h$ & $10^{-1}, 10^{-2}, 10^{-3}, 10^{-4}$ & $10^{-2}$\\
    Confidence parameter (b) & $5.0\cdot 10^{-1}, 10\cdot 10^{-1}, 15\cdot 10^{-1}$ & $5.0\cdot 10^{-1}$\\
  \end{tabular}
\end{table*}

\newpage

\subsection{Ablation Study Difference Classifier Learning Rate} \label{sec:lr}
\begin{samepage}
\begin{figure*}[!htb]
\minipage{0.32\textwidth}
  \includegraphics[width=\linewidth,height=0.9\textwidth]{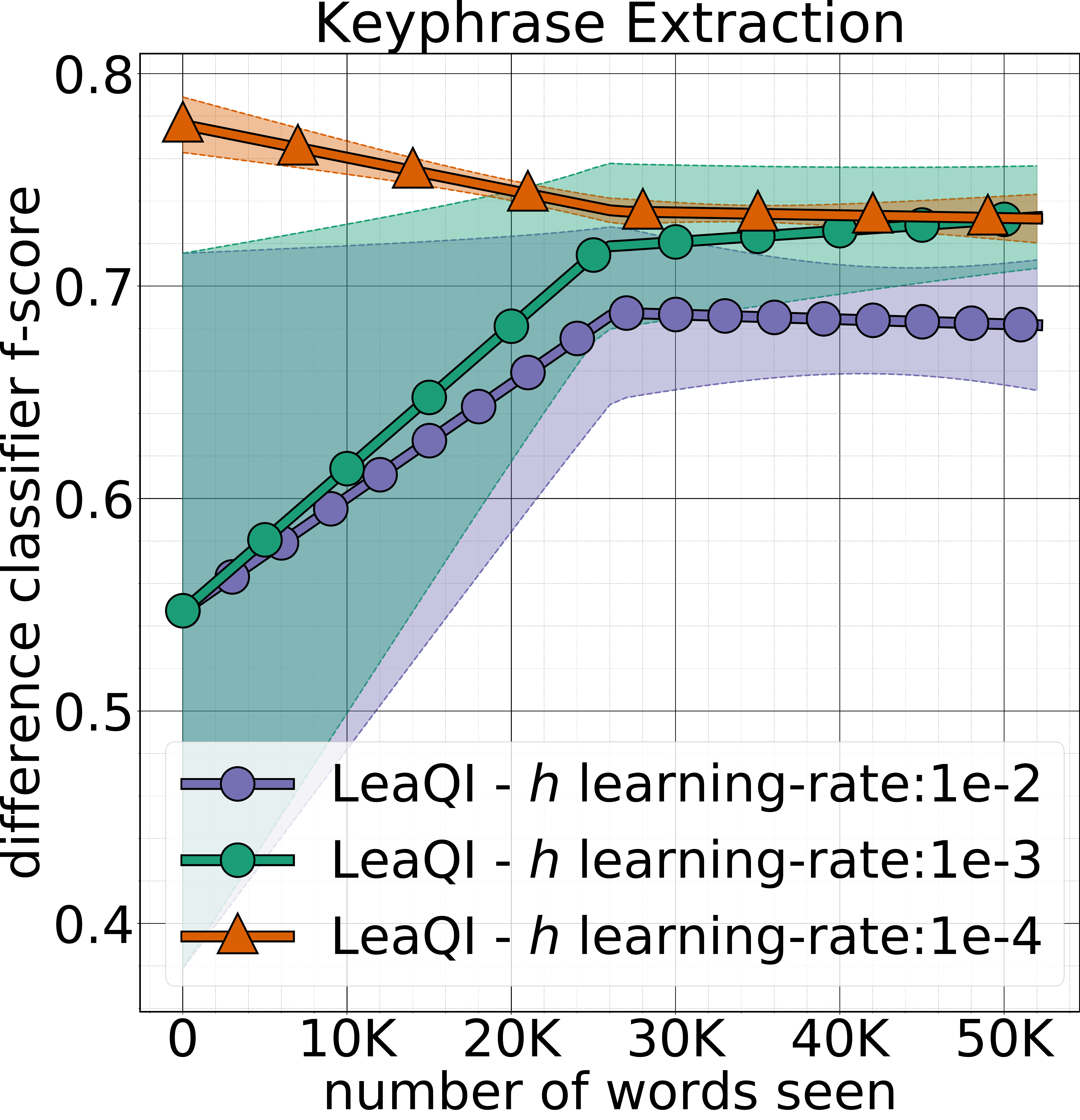}
\endminipage\hfill
\minipage{0.32\textwidth}
  \includegraphics[width=\linewidth,height=0.9\textwidth]{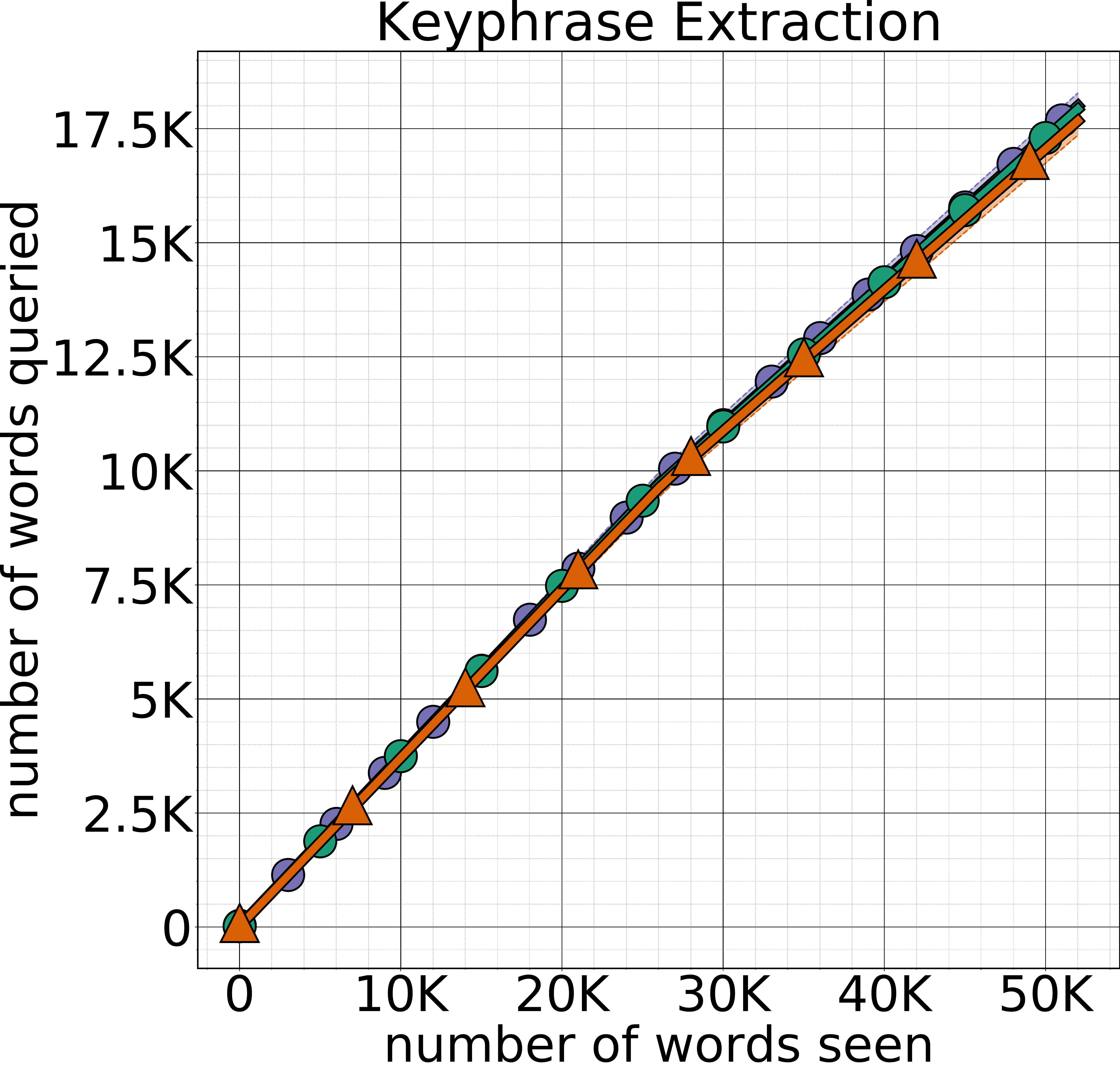}
\endminipage\hfill
\minipage{0.32\textwidth}%
  \includegraphics[width=\linewidth,height=0.9\textwidth]{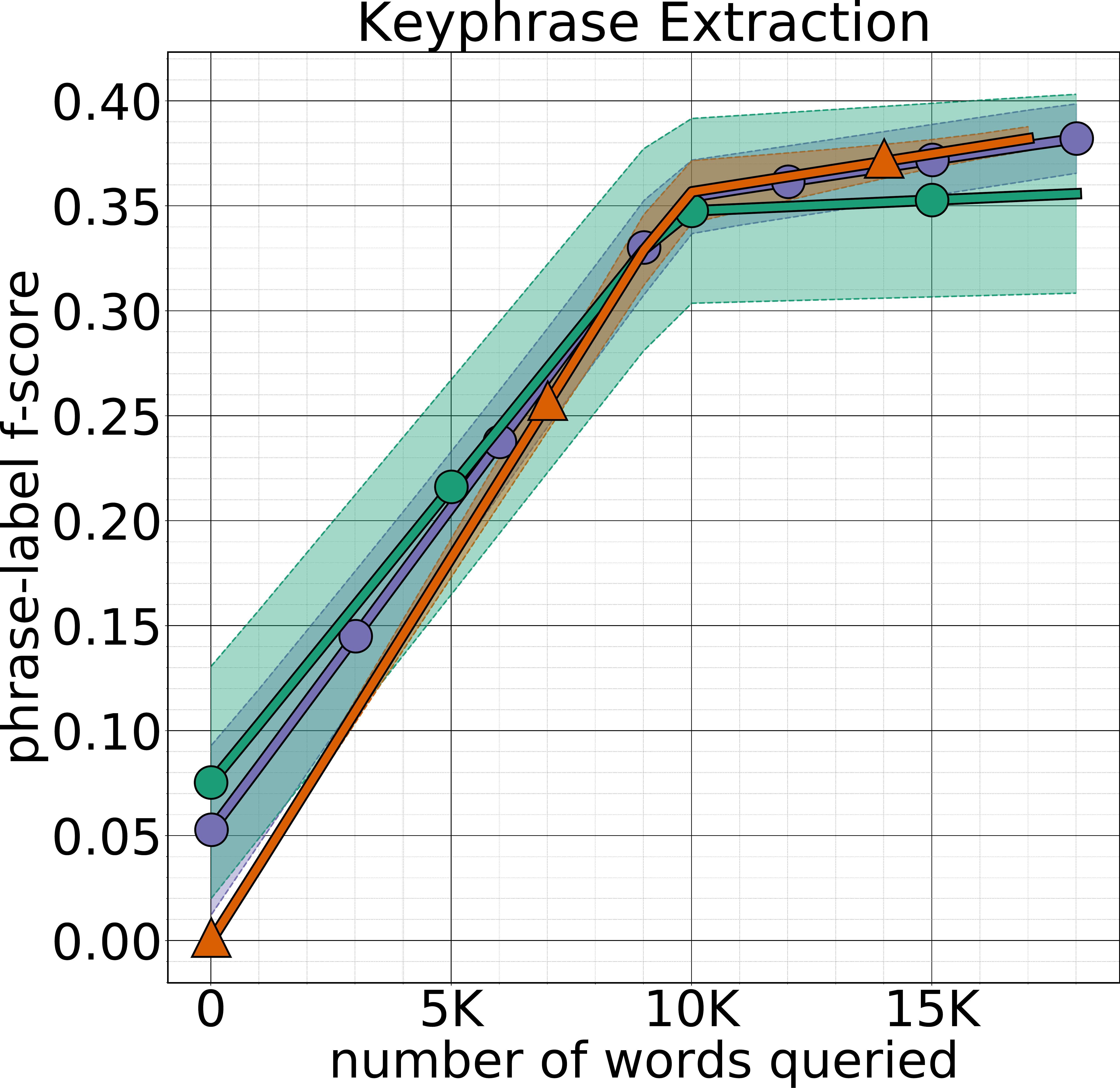}
\endminipage
\end{figure*}
\begin{figure*}[!htb]
\minipage{0.32\textwidth}
  \includegraphics[width=\linewidth,height=0.9\textwidth]{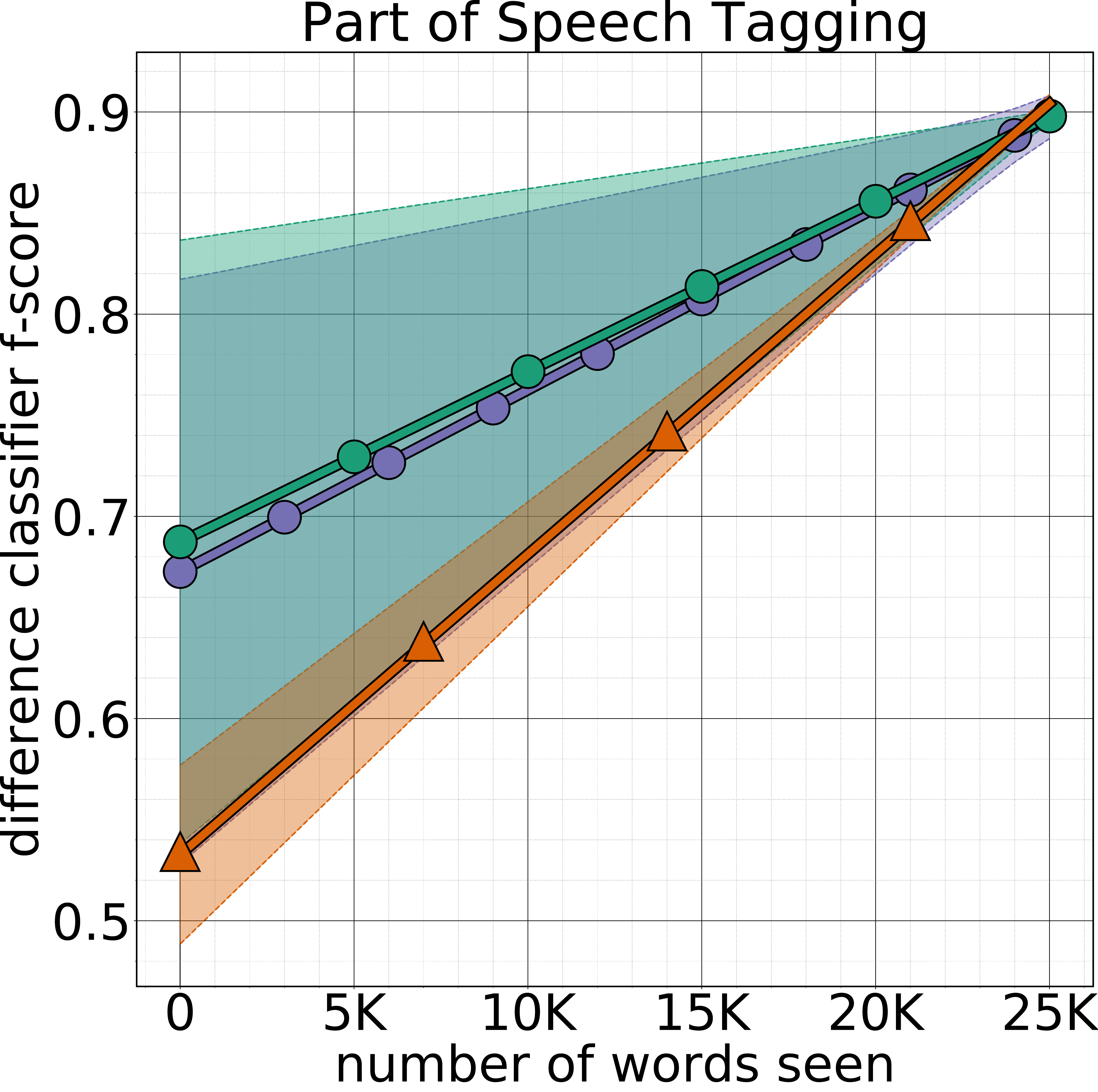}
\endminipage\hfill
\minipage{0.32\textwidth}
  \includegraphics[width=\linewidth,height=0.9\textwidth]{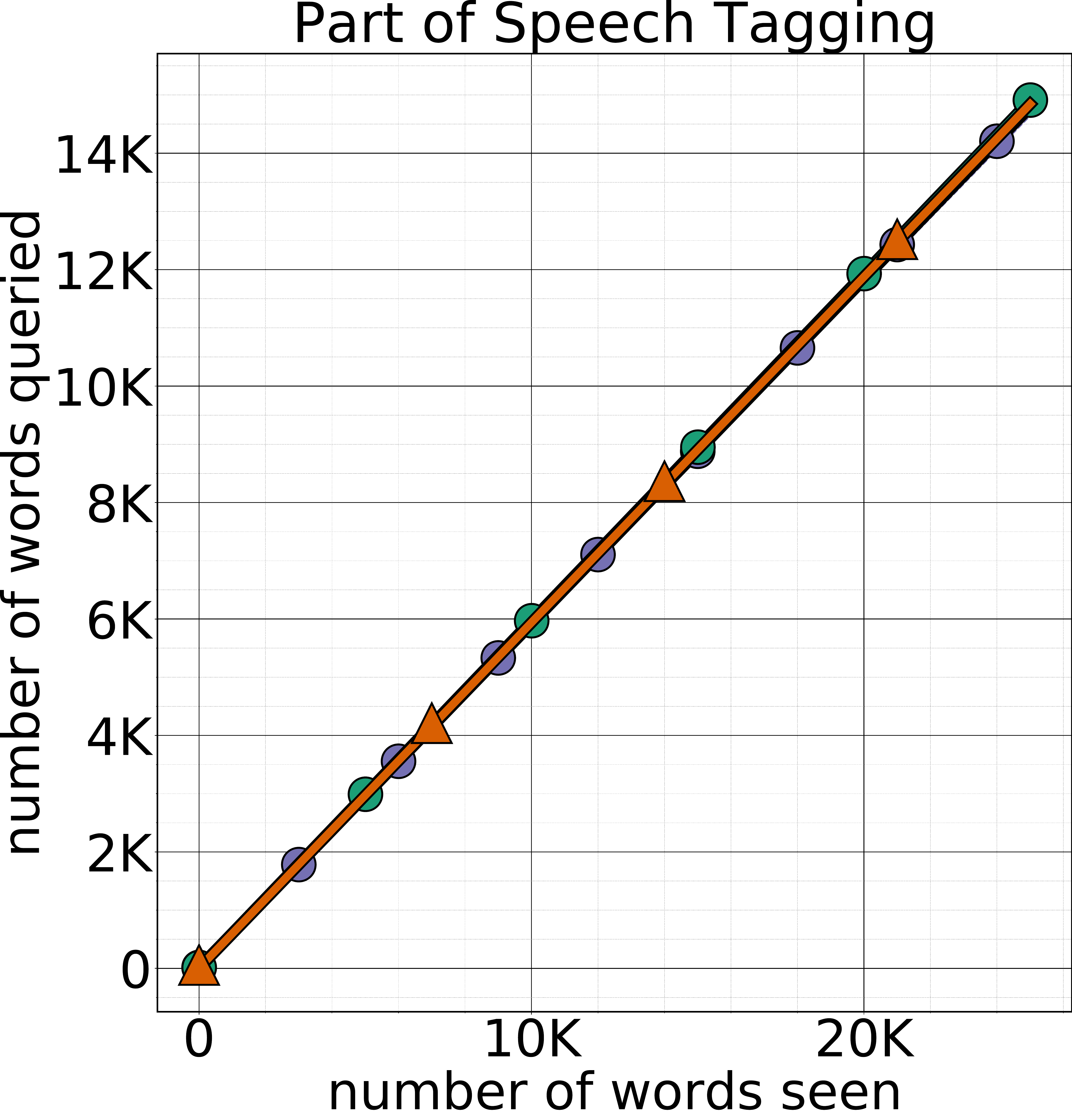}
\endminipage\hfill
\minipage{0.32\textwidth}%
  \includegraphics[width=\linewidth,height=0.9\textwidth]{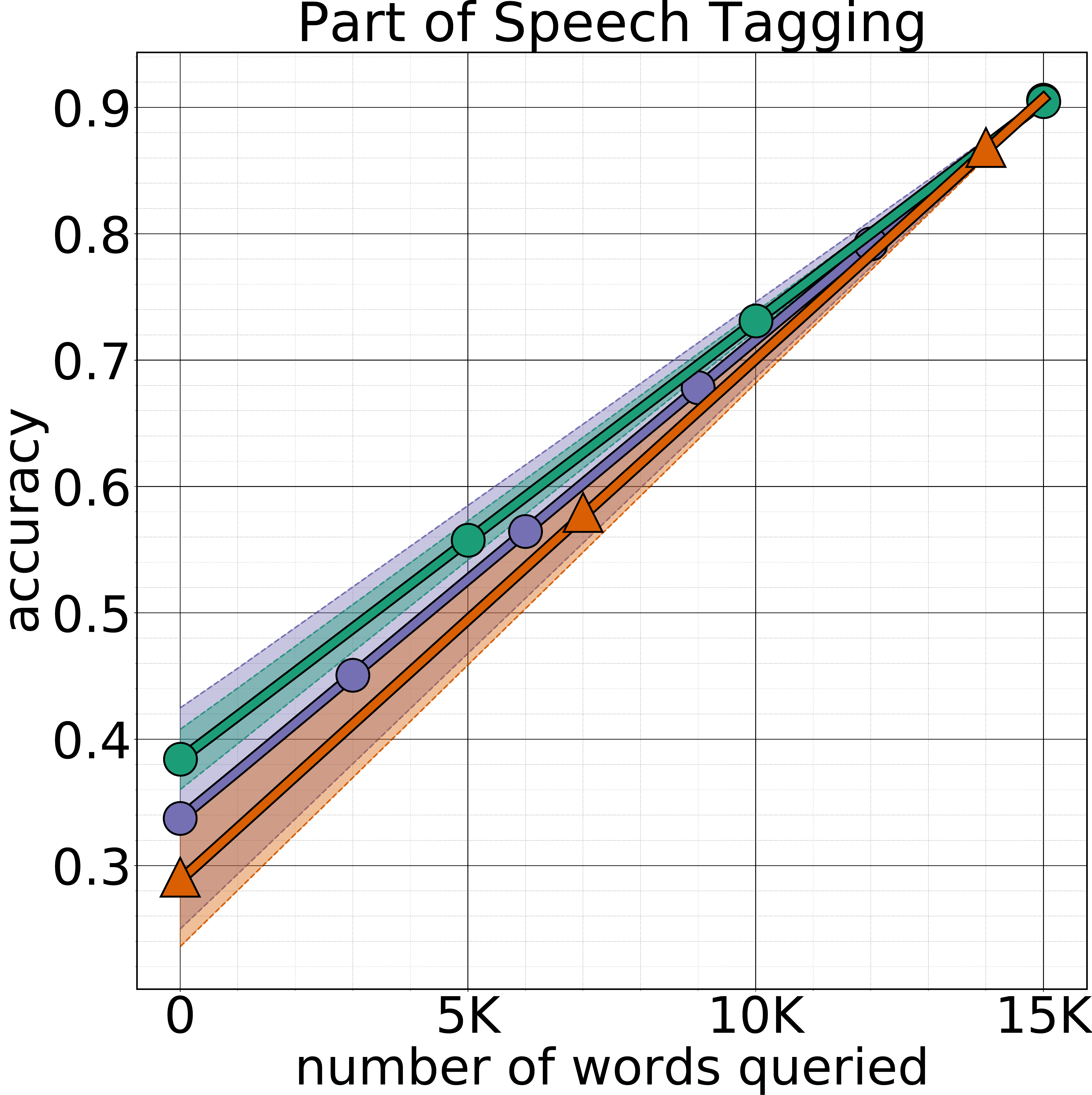}
\endminipage
\caption{(top-row) English keyphrase extraction and (bottom-row) low-resource language part of speech tagging on Greek, Modern (el). We show the performance of using different learning for the difference classifier $h$. These plots indicate that their is small difference in performance depending on the difference classifier learning rate.}
\end{figure*}
\end{samepage}

\newpage

\subsection{Ablation Study Confidence Parameter: $b$} \label{sec:b}
\begin{samepage}
\begin{figure*}[!htb]
\minipage{0.16\textwidth}
\endminipage\hfill
\minipage{0.32\textwidth}
  \includegraphics[width=\linewidth,height=0.9\textwidth]{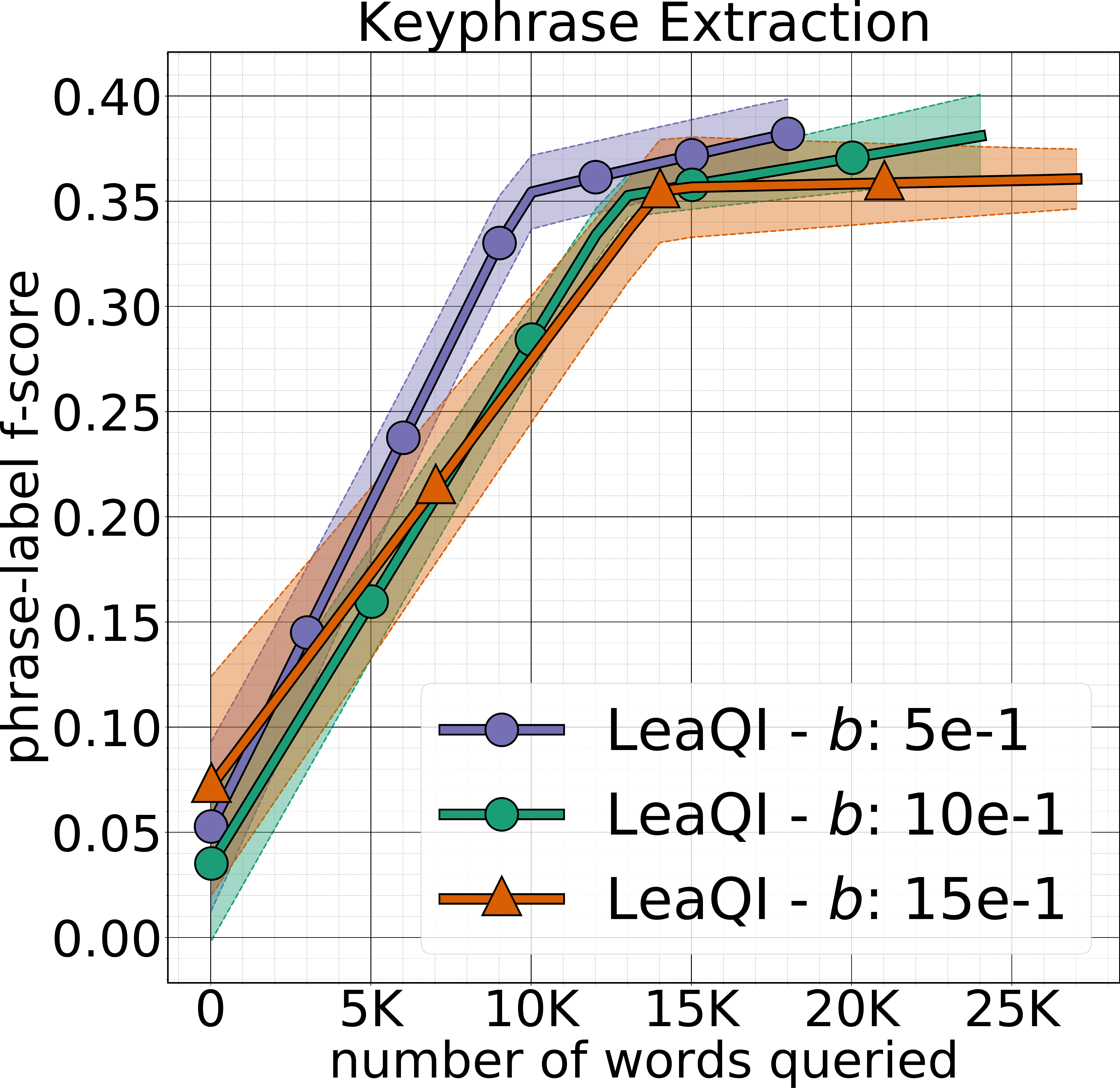}
\endminipage\hfill
\minipage{0.32\textwidth}
  \includegraphics[width=\linewidth,height=0.9\textwidth]{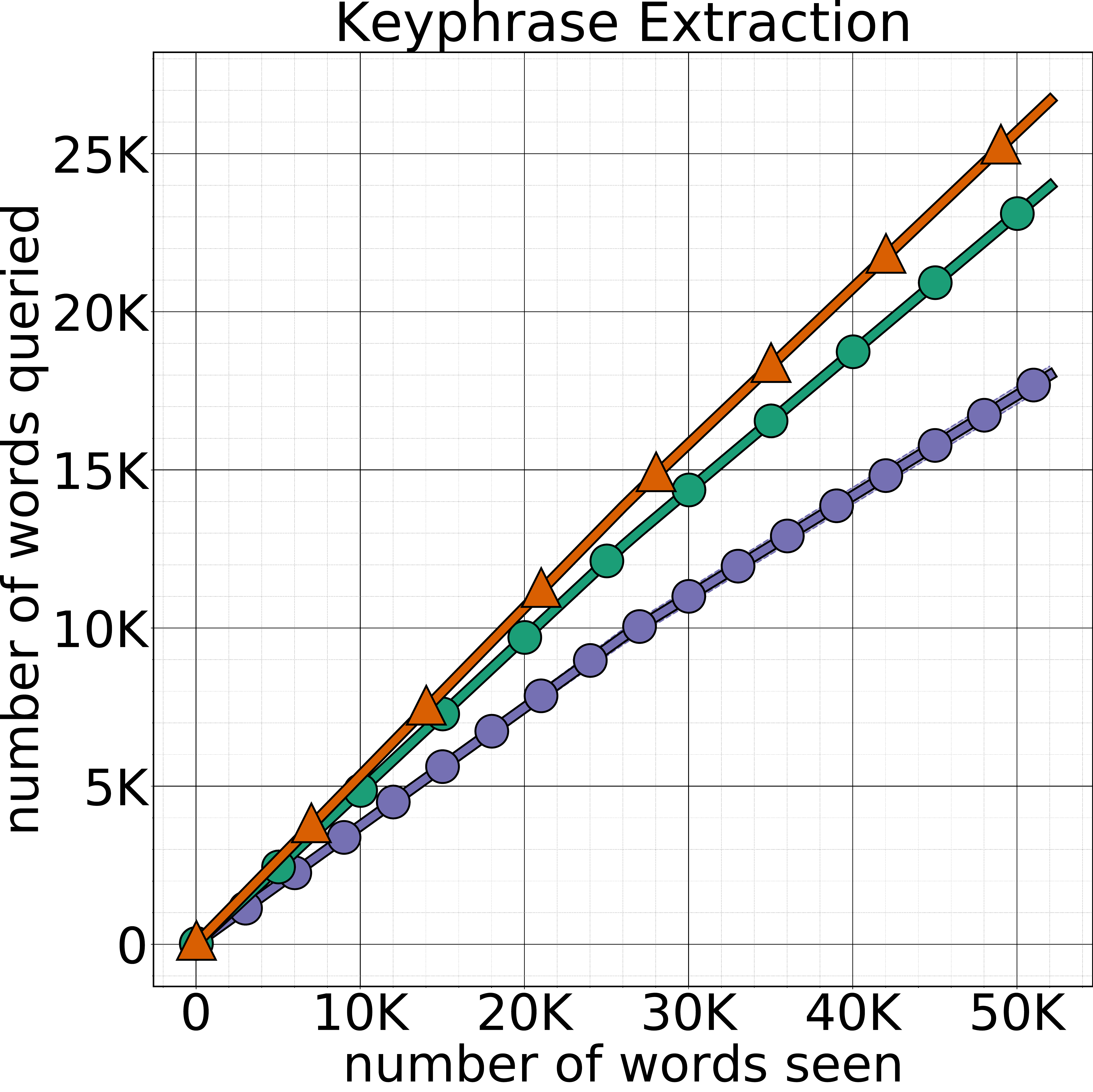}
\endminipage\hfill
\minipage{0.16\textwidth}
\endminipage\hfill
\end{figure*}
\begin{figure*}[!htb]
\minipage{0.16\textwidth}
\endminipage\hfill
\minipage{0.32\textwidth}
  \includegraphics[width=\linewidth,height=0.9\textwidth]{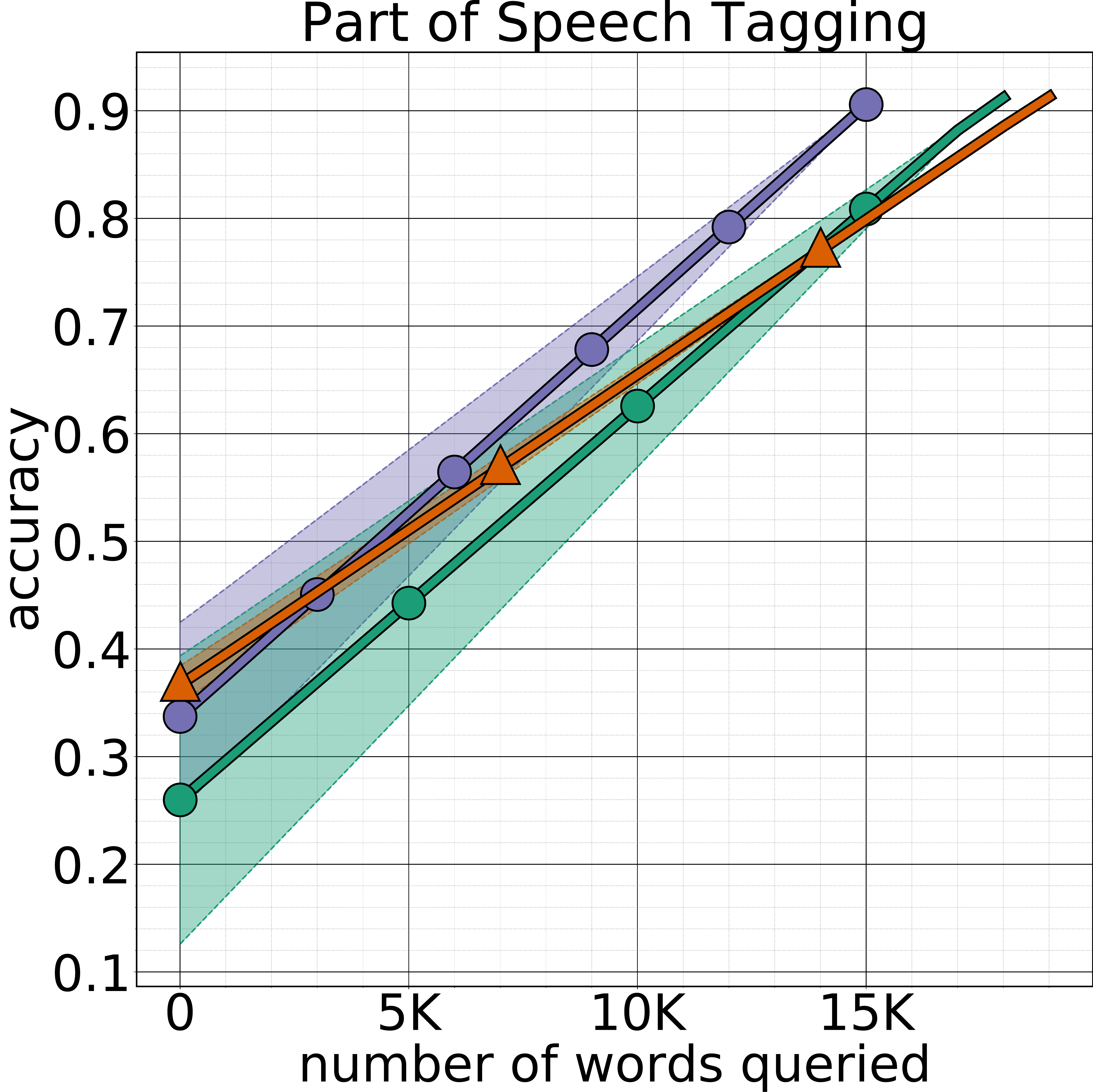}
\endminipage\hfill
\minipage{0.32\textwidth}
  \includegraphics[width=\linewidth,height=0.9\textwidth]{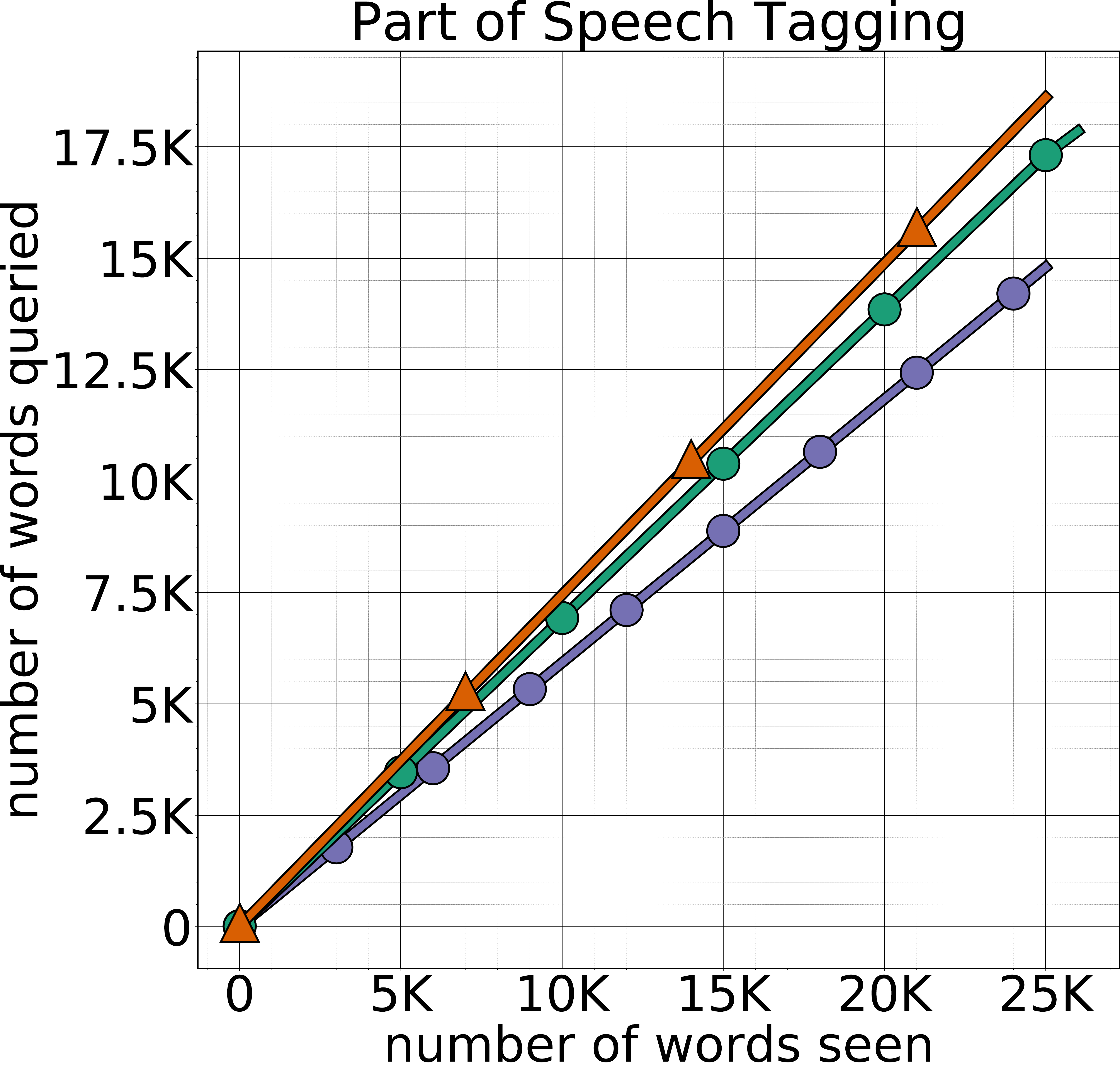}
\endminipage\hfill
\minipage{0.16\textwidth}
\endminipage\hfill
\caption{(top-row) English keyphrase extraction and (bottom-row) low-resource language part of speech tagging on Greek, Modern (el). We show the performance of using difference confidence parameters $b$. These plots indicate that our model is robust to difference confidence parameters.}
\end{figure*}
\end{samepage}

%% file: acl2019.bbl
\begin{thebibliography}{38}
\expandafter\ifx\csname natexlab\endcsname\relax\def\natexlab#1{#1}\fi

\bibitem[{Atlas et~al.(1990)Atlas, Cohn, and Ladner}]{atlas1990training}
Les~E Atlas, David~A Cohn, and Richard~E Ladner. 1990.
\newblock Training connectionist networks with queries and selective sampling.
\newblock In \emph{NeurIPS}.

\bibitem[{Augenstein et~al.(2017)Augenstein, Das, Riedel, Vikraman, and
  McCallum}]{augenstein2017semeval}
Isabelle Augenstein, Mrinal Das, Sebastian Riedel, Lakshmi Vikraman, and Andrew
  McCallum. 2017.
\newblock Semeval 2017 task 10: Scienceie - extracting keyphrases and relations
  from scientific publications.
\newblock In \emph{Proceedings of the 11th International Workshop on Semantic
  Evaluation (SemEval-2017)}.

\bibitem[{Balcan et~al.(2006)Balcan, Beygelzimer, and
  Langford}]{balcan06agnostic}
Nina Balcan, Alina Beygelzimer, and John Langford. 2006.
\newblock Agnostic active learning.
\newblock In \emph{ICML}.

\bibitem[{Beltagy et~al.(2019)Beltagy, Lo, and Cohan}]{Beltagy2019SciBERT}
Iz~Beltagy, Kyle Lo, and Arman Cohan. 2019.
\newblock Scibert: Pretrained language model for scientific text.
\newblock In \emph{EMNLP}.

\bibitem[{Bengio et~al.(2015)Bengio, Vinyals, Jaitly, and
  Shazeer}]{bengio-scheduled}
Samy Bengio, Oriol Vinyals, Navdeep Jaitly, and Noam Shazeer. 2015.
\newblock Scheduled sampling for sequence prediction with recurrent neural
  networks.
\newblock In \emph{NeurIPS}.

\bibitem[{Beygelzimer et~al.(2009)Beygelzimer, Dasgupta, , and
  Langford}]{beygelzimer09active}
Alina Beygelzimer, Sanjoy Dasgupta, , and John Langford. 2009.
\newblock Importance weighted active learning.
\newblock In \emph{ICML}.

\bibitem[{Beygelzimer et~al.(2010)Beygelzimer, Hsu, Langford, and
  Zhang}]{beygelzimer10agnostic}
Alina Beygelzimer, Daniel Hsu, John Langford, and Tong Zhang. 2010.
\newblock Agnostic active learning without constraints.
\newblock In \emph{NeurIPS}.

\bibitem[{Bloodgood and Callison-Burch(2010)}]{bloodgood2010bucking}
Michael Bloodgood and Chris Callison-Burch. 2010.
\newblock Bucking the trend: Large-scale cost-focused active learning for
  statistical machine translation.
\newblock In \emph{ACL}.

\bibitem[{Cesa-Bianchi et~al.(2006)Cesa-Bianchi, Gentile, and
  Zaniboni}]{cesa-bianchi06selective}
Nicol\`o Cesa-Bianchi, Claudio Gentile, and Luca Zaniboni. 2006.
\newblock Worst-case analysis ofselective sampling for linear classification.
\newblock \emph{JMLR}.

\bibitem[{Collins and Roark(2004)}]{collins04incremental}
Michael Collins and Brian Roark. 2004.
\newblock Incremental parsing with the perceptron algorithm.
\newblock In \emph{ACL}.

\bibitem[{Culotta and McCallum(2005)}]{culotta05effort}
Aron Culotta and Andrew McCallum. 2005.
\newblock Reducing labeling effort for structured prediction tasks.
\newblock In \emph{AAAI}.

\bibitem[{Daum\'e et~al.(2009)Daum\'e, Langford, and Marcu}]{daume09searn}
Hal Daum\'e, III, John Langford, and Daniel Marcu. 2009.
\newblock Search-based structured prediction.
\newblock \emph{Machine Learning Journal}.

\bibitem[{Devlin et~al.(2019)Devlin, Chang, Lee, and
  Toutanova}]{devlin2019bert}
Jacob Devlin, Ming-Wei Chang, Kenton Lee, and Kristina Toutanova. 2019.
\newblock {BERT}: Pre-training of deep bidirectional transformers for language
  understanding.
\newblock In \emph{NAACL}.

\bibitem[{Florescu and Caragea(2017)}]{florescu2017positionrank}
Corina Florescu and Cornelia Caragea. 2017.
\newblock {P}osition{R}ank: An unsupervised approach to keyphrase extraction
  from scholarly documents.
\newblock In \emph{ACL}.

\bibitem[{Hachey et~al.(2005)Hachey, Alex, and
  Becker}]{hachey2005investigating}
Ben Hachey, Beatrice Alex, and Markus Becker. 2005.
\newblock Investigating the effects of selective sampling on the annotation
  task.
\newblock In \emph{CoNLL}.

\bibitem[{Haertel et~al.(2008)Haertel, Ringger, Seppi, Carroll, and
  McClanahan}]{haertel2008assessing}
Robbie Haertel, Eric~K. Ringger, Kevin~D. Seppi, James~L. Carroll, and Peter
  McClanahan. 2008.
\newblock Assessing the costs of sampling methods in active learning for
  annotation.
\newblock In \emph{ACL}.

\bibitem[{Haghighi and Klein(2006)}]{Haghighi2006}
Aria Haghighi and Dan Klein. 2006.
\newblock Prototype-driven learning for sequence models.

\bibitem[{Helmbold et~al.(2000)Helmbold, Littlestone, and Long}]{Helmbold2000}
David~P. Helmbold, Nicholas Littlestone, and Philip~M. Long. 2000.
\newblock Apple tasting.
\newblock \emph{Information and Computation}.

\bibitem[{Judah et~al.(2012)Judah, Fern, and Dietterich}]{judah2012active}
Kshitij Judah, Alan~Paul Fern, and Thomas~Glenn Dietterich. 2012.
\newblock Active imitation learning via reduction to iid active learning.
\newblock In \emph{AAAI}.

\bibitem[{Khashabi et~al.(2018)Khashabi, Sammons, Zhou, Redman,
  Christodoulopoulos, Srikumar, Rizzolo, Ratinov, Luo, Do, Tsai, Roy, Mayhew,
  Feng, Wieting, Yu, Song, Gupta, Upadhyay, Arivazhagan, Ning, Ling, and
  Roth}]{cogcompnlp2018}
Daniel Khashabi, Mark Sammons, Ben Zhou, Tom Redman, Christos
  Christodoulopoulos, Vivek Srikumar, Nicholas Rizzolo, Lev Ratinov, Guanheng
  Luo, Quang Do, Chen-Tse Tsai, Subhro Roy, Stephen Mayhew, Zhili Feng, John
  Wieting, Xiaodong Yu, Yangqiu Song, Shashank Gupta, Shyam Upadhyay, Naveen
  Arivazhagan, Qiang Ning, Shaoshi Ling, and Dan Roth. 2018.
\newblock {CogCompNLP}: Your swiss army knife for {NLP}.
\newblock In \emph{LREC}.

\bibitem[{Leblond et~al.(2018)Leblond, Alayrac, Osokin, and
  Lacoste-Julien}]{leblond18searnn}
R\'emi Leblond, Jean-Baptiste Alayrac, Anton Osokin, and Simon Lacoste-Julien.
  2018.
\newblock {SEARNN}: Training {RNN}s with global-local losses.
\newblock In \emph{ICLR}.

\bibitem[{Lee et~al.(2011)Lee, Peirsman, Chang, Chambers, Surdeanu, and
  Jurafsky}]{stanford-coref}
Heeyoung Lee, Yves Peirsman, Angel Chang, Nathanael Chambers, Mihai Surdeanu,
  and Dan Jurafsky. 2011.
\newblock Stanford's multi-pass sieve coreference resolution system at the
  conll-2011 shared task.
\newblock In \emph{Proceedings of the Fifteenth Conference on Computational
  Natural Language Learning: Shared Task}.

\bibitem[{Littlestone and Warmuth(1989)}]{Littlestone1989WMA}
N.~Littlestone and M.~K. Warmuth. 1989.
\newblock The weighted majority algorithm.
\newblock In \emph{Proceedings of the 30th Annual Symposium on Foundations of
  Computer Science}.

\bibitem[{Nivre(2018)}]{Nivre2018UD}
Joakim et.~al Nivre. 2018.
\newblock Universal dependencies v2.5.
\newblock LINDAT/CLARIN digital library at the Institute of Formal and Applied
  Linguistics, Charles University.

\bibitem[{Ratnaparkhi(1996)}]{ratnaparkhi1996maximum}
Adwait Ratnaparkhi. 1996.
\newblock A maximum entropy model for part-of-speech tagging.
\newblock In \emph{EMNLP}.

\bibitem[{Rendell(1986)}]{rendell1986general}
Larry Rendell. 1986.
\newblock A general framework for induction and a study of selective induction.
\newblock \emph{Machine Learning Journal}.

\bibitem[{Riloff and Wiebe(2003)}]{riloff2003learning}
Ellen Riloff and Janyce Wiebe. 2003.
\newblock Learning extraction patterns for subjective expressions.
\newblock In \emph{EMNLP}.

\bibitem[{Ringger et~al.(2007)Ringger, McClanahan, Haertel, Busby, Carmen,
  Carroll, Seppi, and Lonsdale}]{Ringger2007ActiveLF}
Eric Ringger, Peter McClanahan, Robbie Haertel, George Busby, Marc Carmen,
  James Carroll, Kevin Seppi, and Deryle Lonsdale. 2007.
\newblock Active learning for part-of-speech tagging: Accelerating corpus
  annotation.
\newblock In \emph{Proceedings of the Linguistic Annotation Workshop}.

\bibitem[{Ross et~al.(2011)Ross, Gordon, and Bagnell}]{ross11dagger}
St\'ephane Ross, Geoff~J. Gordon, and J.~Andrew Bagnell. 2011.
\newblock A reduction of imitation learning and structured prediction to
  no-regret online learning.
\newblock In \emph{AI-Stats}.

\bibitem[{Sculley(2007)}]{sculley2007}
David Sculley. 2007.
\newblock Practical learning from one-sided feedback.
\newblock In \emph{KDD}.

\bibitem[{Singh(2017)}]{Singh2017}
Vikash Singh. 2017.
\newblock Replace or retrieve keywords in documents at scale.
\newblock \emph{CoRR}, abs/1711.00046.

\bibitem[{Smit et~al.(2014)Smit, Virpioja, Gr{\"o}nroos, and
  Kurimo}]{smit2014morfessor}
Peter Smit, Sami Virpioja, Stig-Arne Gr{\"o}nroos, and Mikko Kurimo. 2014.
\newblock {M}orfessor 2.0: Toolkit for statistical morphological segmentation.
\newblock In \emph{EACL}.

\bibitem[{Thompson et~al.(1999)Thompson, Califf, and
  Mooney}]{thompson99parsing}
Cynthia~A. Thompson, Mary~Elaine Califf, and Raymond~J. Mooney. 1999.
\newblock Active learning for natural language parsing and information
  extraction.
\newblock In \emph{ICML}.

\bibitem[{Tjong Kim~Sang and De~Meulder(2003)}]{TjongKimSang2003}
Erik~F. Tjong Kim~Sang and Fien De~Meulder. 2003.
\newblock Introduction to the {CoNLL}-2003 shared task: Language-independent
  named entity recognition.
\newblock In \emph{NAACL/HLT}.

\bibitem[{Vapnik(1982)}]{vapnik1982}
Vladimir Vapnik. 1982.
\newblock \emph{Estimation of Dependences Based on Empirical Data: Springer
  Series in Statistics (Springer Series in Statistics)}.
\newblock Springer-Verlag, Berlin, Heidelberg.

\bibitem[{Whitehead(1991)}]{whitehead1991study}
Steven Whitehead. 1991.
\newblock A study of cooperative mechanisms for faster reinforcement learning.
\newblock Technical report, University of Rochester.

\bibitem[{Zesch et~al.(2008)Zesch, M{\"u}ller, and
  Gurevych}]{zesch2008extracting}
Torsten Zesch, Christof M{\"u}ller, and Iryna Gurevych. 2008.
\newblock Extracting lexical semantic knowledge from {W}ikipedia and
  {W}iktionary.
\newblock In \emph{LREC}.

\bibitem[{Zhang and Chaudhuri(2015)}]{zhang2015active}
Chicheng Zhang and Kamalika Chaudhuri. 2015.
\newblock Active learning from weak and strong labelers.
\newblock In \emph{NeurIPS}.

\end{thebibliography}
